\PassOptionsToPackage{table,dvipsnames}{xcolor}
\documentclass[]{selfevolagent}
\usepackage{microtype}
\usepackage{amsfonts}
\usepackage{xcolor}
\definecolor{lavender}{RGB}{230,230,250} 
\definecolor{softlavender}{RGB}{238, 223, 255} 
\definecolor{selfevolagent}{RGB}{220,211,237} 
\usepackage{graphicx}
\usepackage{booktabs}
\usepackage{wrapfig}
\usepackage{float}
\usepackage{amsmath}
\usepackage{amsthm}
\usepackage{subcaption}
\usepackage{makecell}
\usepackage{algpseudocode}
\usepackage[linesnumbered,lined,boxed,commentsnumbered,ruled,longend]{algorithm2e}
\usepackage[capitalize,noabbrev]{cleveref}
\theoremstyle{plain}

\theoremstyle{definition}

\theoremstyle{remark}

\usepackage{enumitem}
\usepackage{pifont}
\usepackage[edges]{forest}
\usetikzlibrary{arrows.meta}
\usepackage{tikz}
\usepackage{graphicx}     
\usepackage{booktabs}     
\usepackage{enumitem}     
\usepackage[most]{tcolorbox} 
\usepackage{fontawesome5}
\usepackage{booktabs} 
\usepackage{url}
\usepackage{wrapfig}
\usepackage{longtable}
\usetikzlibrary{
  positioning,
  calc,
  shapes.symbols,
  shapes.geometric,
  shapes.misc
}
\usepackage[T1]{fontenc}
\definecolor{selfevolagent_dark}{HTML}{37D2A6} 
\definecolor{selfevolagent_light}{HTML}{9BE9D3}
\definecolor{selfevolagent_lighter}{HTML}{CDF4E9}
\newcommand{\ThreeLaws}{{\scshape\color{selfevolagent_dark!120} Three Laws of Self-Evolving AI Agents}}

\title{A Comprehensive Survey of Self-Evolving AI Agents\\
\Large A New Paradigm Bridging Foundation Models and Lifelong Agentic Systems}
\author{Jinyuan Fang$^{*1}$}
\author{Yanwen Peng$^{*2}$}
\author{Xi Zhang$^{*1}$}
\author{Yingxu Wang$^{3}$}
\author{Xinhao Yi$^1$, Guibin Zhang$^4$, Yi Xu$^5$, Bin Wu$^6$, Siwei Liu$^7$, Zihao Li$^1$, Zhaochun Ren$^8$, Nikos Aletras$^2$, Xi Wang$^2$, Han Zhou$^5$, Zaiqiao Meng${^1}${\large \ding{41}}}
\affiliation{$^1$University of Glasgow, $^2$University of Sheffield, $^3$Mohamed bin Zayed University of Artificial Intelligence, $^4$National University of Singapore,  $^5$University of Cambridge, $^6$University College London, $^7$University of Aberdeen, $^8$Leiden University}

\contribution[*]{Equal Contributor}
\contribution[\text{\large \ding{41}}]{Corresponding Author}

\abstract{
Recent advances in large language models (LLMs) have sparked growing interest in AI agents capable of solving complex, real-world tasks. However, most existing agent systems rely on manually crafted configurations that remain static after deployment, limiting their ability to adapt to dynamic and evolving environments. To address this limitation, recent research has explored agent \textit{evolution} techniques that aim to automatically enhance agent systems based on interaction data and environmental feedback. This emerging direction lays the foundation for \textit{self-evolving AI agents}, which bridge the static capabilities of foundation models with the continuous adaptability required by \textit{lifelong agentic systems}. In this survey, we provide a comprehensive review of existing techniques for self-evolving agentic systems. Specifically, we first introduce a \textit{unified conceptual framework} that abstracts the feedback loop underlying the design of self-evolving agentic systems. The framework highlights four key components: \textit{System inputs}, \textit{Agent System}, \textit{Environment}, and \textit{Optimisers}, serving as a foundation for understanding and comparing different strategies. Based on this framework, we systematically review a wide range of self-evolving techniques that target different components of the agent system, including foundation models, agent prompts, memory, tools, workflows, and communication mechanisms across agents. We also investigate domain-specific evolution strategies developed for specialised fields such as biomedicine, programming, and finance, where agent behaviour and optimisation objectives are tightly coupled with domain constraints. In addition, we provide a dedicated discussion on the \textit{evaluation, safety, and ethical considerations} for self-evolving agentic systems, which are critical to ensuring their effectiveness and reliability. This survey aims to provide researchers and practitioners with a systematic understanding of self-evolving AI agents, laying the foundation for the development of more adaptive, autonomous, and lifelong agentic systems. 
}
\metadata[{\raisebox{-0.2ex}{\includegraphics[height=1em]{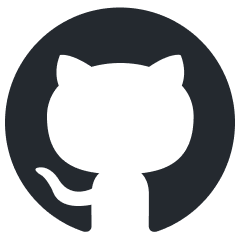}}\ Github}]{\url{https://github.com/EvoAgentX/Awesome-Self-Evolving-Agents}}

\begin{document}

\maketitle

\begin{figure}
    \centering
    \includegraphics[width=0.9\linewidth]{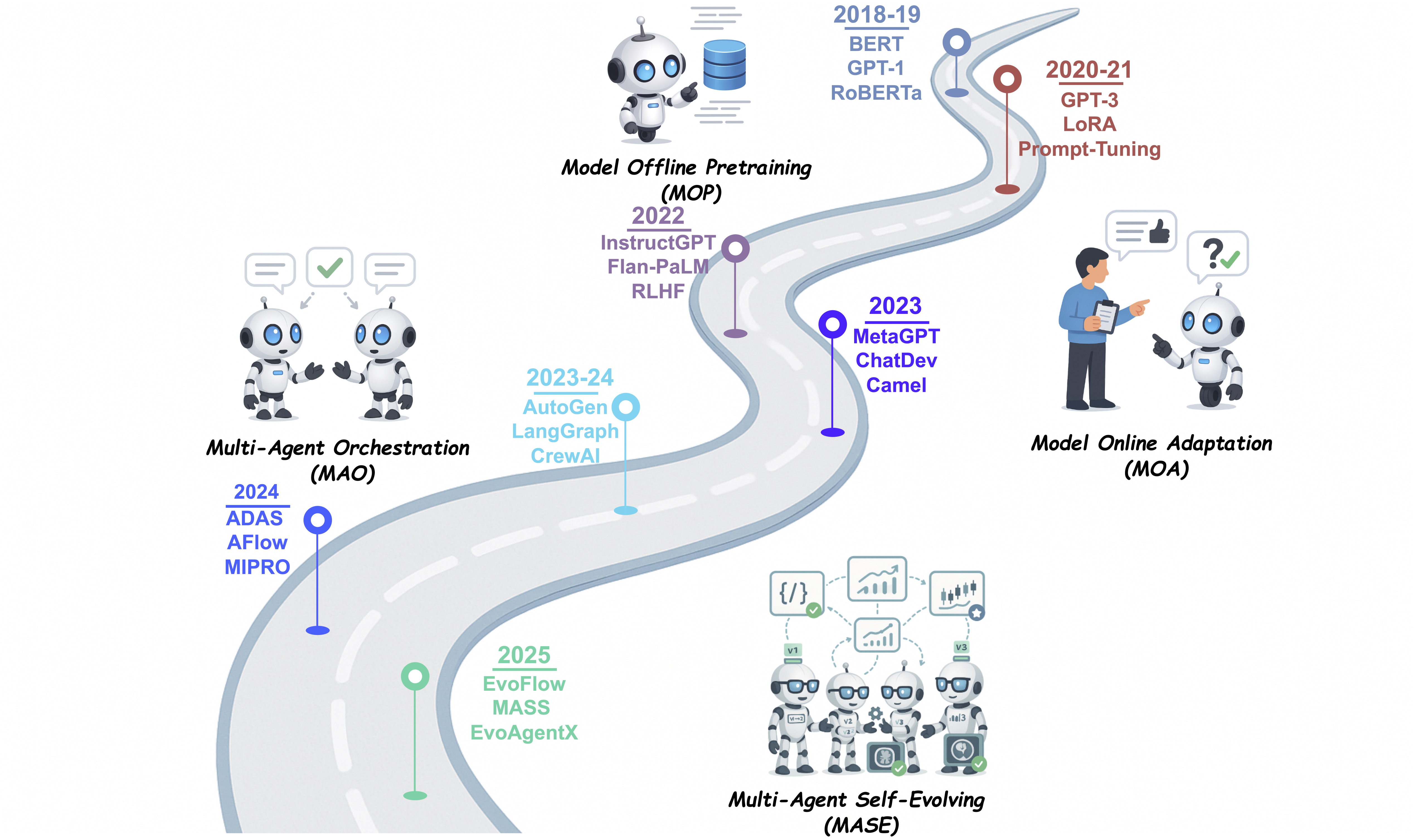}
    \caption{LLM-centric learning is evolving from learning purely from static data, to interacting with dynamic environments, and ultimately towards lifelong learning through multi-agent collaboration and self-evolution.}
    \label{fig:path}
\end{figure}

\section{Introduction}

Recent progress in large language models (LLMs) has significantly advanced the development of artificial intelligence (AI). Owing to progress in large-scale pretraining, supervised fine-tuning and reinforcement learning, LLMs have demonstrated remarkable capabilities in planning, reasoning, and natural language understanding~\citep{zhao2023survey,dubey2024llama,yang2025qwen3,guo2025deepseek}. These advances have sparked growing interest in \textit{LLM-based agents} (a subclass of AI agents in which an LLM serves as the decision/policy module)~\citep{wang2024survey,luo2025large}, which are \textit{autonomous systems that leverage LLMs as the core reasoning components for understanding inputs, planning actions, and generating outputs in open-ended, real-world environments}~\citep{wang2024survey,xi2025the,luo2025large}. A typical AI agent consists of several components that enable it to perform complex, goal-oriented tasks in an autonomous manner. The foundation model (e.g. an LLM) is the core, responsible for interpreting goals, making plans, and executing actions. To support these capabilities, additional modules, such as perception~\citep{shridhar2021alfworld,zheng2024steve}, planning~\citep{yao2023tree,yao2023react,besta2024graph}, memory~\citep{modarressi2023ret,zhong2023memorybankenhancinglargelanguage}, and tools~\citep{schick2023toolformer,tora,liu2025toolace}, are integrated to help the agent perceive inputs, decompose tasks, retain contextual information, and interact with tools~\citep{wang2024survey}. 

While single-agent systems have demonstrated strong generalisation and adaptability in various tasks, they often struggle with task specialisation and coordination in dynamic and complex environments~\citep{wu2024autogen,qian2024chatdevcommunicativeagentssoftware}. These limitations have led to the development of multi-agent systems~(MAS)~\citep{hong2023metagpt,guo2024large,zhou2025multi}, where multiple agents collaborate to solve complex problems. Compared with single-agent systems, MAS enables functional specialisation, with each agent designed for a specific subtask or domain of expertise. 
Moreover, agents can interact, exchange information, and coordinate their behaviour to achieve shared goals. Such collaboration enables the system to tackle tasks beyond the capability of a single agent, while simulating more realistic, dynamic, and interactive environments. LLM-based agent systems have been successfully applied to a wide range of real-world tasks, ranging from code generation~\citep{jiang2024survey}, scientific research~\citep{lu2024ai}, web navigation~\citep{lai2024autowebglm}, to domain-specific applications in biomedicine~\citep{kim2024mdagents} and finance~\citep{tian2025template}. 

Despite the notable progress in agent systems, most of them, whether single- or multi-agent, continue to rely extensively on manually designed configurations. Once deployed, these systems typically maintain static architectures and fixed functionalities.
However, real-world environments are dynamic and continuously evolving, e.g., user intents shift, task requirements change, and external tools or information sources may vary over time. 
For instance, an agent assisting in customer service may need to handle newly introduced products, updated company policies, or unfamiliar user intents. Similarly, a scientific research assistant may be required to incorporate a newly published algorithm, or integrate a novel analysis tool. 
In such settings, manually reconfiguring the agent system is time-consuming, labour-intensive, and difficult to scale. 

These challenges have motivated recent efforts to explore the new paradigm of \textit{\textbf{Self-Evolving AI Agents}}, a novel class of agent systems capable of autonomous adaptation and continuous self-improvement, bridging foundation models with lifelong learning agentic systems.

\vspace{0.6em}
\begin{tcolorbox}[
  colback=selfevolagent_light!20,
  colframe=selfevolagent_light!80,
  colbacktitle=selfevolagent_light!80,
  coltitle=black,
  title={\bfseries\fontfamily{ppl}\selectfont{Definition}},
  boxrule=2pt,
  arc=5pt,
  drop shadow,
  parbox=false,
  before skip=5pt,
  after skip=5pt,
  left=5pt,   
  right=5pt,
]
Self-evolving AI agents are autonomous systems that continuously and systematically optimise their internal components through interaction with environments, with the goal of adapting to changing tasks, contexts and resources while preserving safety and enhancing performance.
\end{tcolorbox}

Inspired by Isaac Asimov’s \textit{Three Laws of Robotics\footnote{Introduced in his stories ``Runaround'' (1942) and ``I, Robot'' (1950). These laws are hierarchical: the Second cannot override the First, and the Third cannot override the First or Second. Although conceived as fictional moral constraints, they have become influential in AI ethics research. Therefore, we articulate the ``Three Laws of Self-Evolving AI Agents'', advocating that AI agents, as the core of embodied AI, prioritise compliance and safety before pursuing autonomous evolution.}}, we propose a set of guiding principles for safe and effective self-evolution of AI agents: 

\vspace{0.6em}
\begin{tcolorbox}[
  colback=selfevolagent_light!20,
  colframe=selfevolagent_light!80,
  colbacktitle=selfevolagent_light!80,
  coltitle=black,
  title={\bfseries\fontfamily{ppl}\selectfont{Three Laws of Self-Evolving AI Agents}},
  boxrule=2pt,
  arc=5pt,
  drop shadow,
  parbox=false,
  before skip=5pt,
  after skip=5pt,
  left=5pt,   
  right=5pt,
]
\begin{enumerate}[label=\Roman*.]
\item \textit{Endure (Safety Adaptation)} \\ Self-evolving AI agents must maintain safety and stability during any modification;
\item \textit{Excel (Performance Preservation)} \\ Subject to the First law, self-evolving AI agents must preserve or enhance existing task performance;
\item \textit{Evolve (Autonomous Evolution)} \\ Subject to the First and Second law, self-evolving AI agents must be able to autonomously optimise their internal components in response to changing tasks, environments, or resources.
\end{enumerate}

\end{tcolorbox}

\vspace{1em} We characterise the emergence of self-evolving AI agents as part of a broader paradigm shift in the development of LLM-based systems. This shift spans from early-stage Model Offline Pretraining (MOP) and Model Online Adaptation (MOA), to more recent trends in Multi-Agent Orchestration (MAO), and ultimately, to Multi-Agent Self-Evolving (MASE). As summarised in Figure \ref{fig:path} and Table~\ref{tab:learning-paradigms}, each paradigm builds on the previous one, moving from a static, frozen foundation model to fully autonomous, self-evolving agentic systems. 
\begin{itemize}
    \item \textbf{MOP (Model Offline Pretraining)}. The initial stage focuses on pretraining foundation models on large-scale, static corpora and then deploying them in a fixed, frozen state, without further adaptation. 
    \item \textbf{MOA (Model Online Adaptation)}. Building on MOP, this stage introduces post-deployment adaptation, where the foundation models can be updated through techniques such as supervised fine-tuning, low-rank adapters~\citep{pfeiffer2020adapterfusion,hu2022lora}, or reinforcement learning from human feedback (RLHF)~\citep{ouyang2022training}, using labels, ratings, or instruction prompts. 
    \item \textbf{MAO (Multi-Agent Orchestration)}. Extending beyond a single foundation model, this stage coordinates multiple LLM agents that communicate and collaborate via message exchange or debate prompts~\citep{li2024improvingmultiagentdebatesparse,zhang2025if}, to solve complex tasks without modifying the underlying model parameters.
    \item \textbf{MASE (Multi-Agent Self-Evolving)}. Finally, MASE introduces a lifelong, self-evolving loop where a population of agents continually refines their prompts, memory, tool-use strategies and even their interaction patterns based on environmental feedback and meta-rewards~\citep{novikov2025alphaevolve,zhang2025darwin}.
\end{itemize}

The evolution from MOP to MASE represents a fundamental shift in the development of LLM-based systems, from static, manually configured architectures to adaptive, data-driven systems that can evolve in response to changing requirements and environments. \textit{Self-evolving AI agents} bridge the static capabilities of foundation models with the continuous adaptability required by \textit{lifelong agentic systems}, offering a path toward more autonomous, resilient, and sustainable AI. 

\begin{table}[t]
\centering
\small
\resizebox{\textwidth}{!}{%
\begin{tabular}{
   >{\centering\arraybackslash}m{3.2cm}
   >{\centering\arraybackslash}m{3.5cm}
   >{\centering\arraybackslash}m{5.0cm}
   >{\centering\arraybackslash}m{4cm}
}
\toprule
\textbf{Paradigm} & \textbf{Interaction \& Feedback} & \textbf{Key Techniques} & \textbf{Diagram} \\
\midrule

\vspace{1.3em}\textbf{Model Offline Pretraining (MOP)}
& 
\vspace{1.3em}
\shortstack[c]{
    {\color{green!50}Model} $\Leftrightarrow$ {\color{orange!50}Static data} \\
    (loss/backprop)
  }
&
\begin{minipage}[t]{\linewidth}
\vspace{-2em}
    \begin{itemize}[nosep,leftmargin=*]
      \item Transformer Pretraining (Causal LM, Masked LM, NSP)
      \item BPE / SentencePiece
      \item MoE \& Pipeline Parallelism
    \end{itemize}
  \end{minipage}
&
\tikz[baseline=(D.base),node distance=7mm]{%
  \node[cylinder,shape border rotate=90,draw,fill=orange!30,minimum height=6mm,minimum width=4mm] (D) {};
  \node[below=1pt of D,font=\tiny] {Static data};
  \node[rectangle,draw,fill=green!30,right=of D,minimum height=6mm,minimum width=8mm] (M) {Model};
  \draw[->,thick] (D.east) -- node[midway,above,font=\tiny]{loss} (M.west);
} \\

\midrule
\textbf{Model Online Adaptation (MOA)}
& \shortstack[c]{
    {\color{green!50}Model} $\Leftrightarrow$ {\color{blue!50}Supervision}\\
    (labels/scores/rewards)
  }
& 
\begin{minipage}[t]{\linewidth}
\vspace{-4em}
    \begin{itemize}[nosep,topsep=0pt,leftmargin=*]
      \item Task Fine-tuning
      \item Instruction Tuning
      \item LoRA / Adapters / Prefix-Tuning
      \item RLHF (RLAIF, DPO, PPO)
      \item Multi-Modal Alignment
      \item Human Alignment
    \end{itemize}
  \end{minipage}
&
\tikz[baseline=(M.base),node distance=8mm and 14mm]{
  \node[rectangle, draw, fill=green!30] (M) {Model};

  \node[rectangle, draw, fill=Aquamarine!10, above right=of M] (MA) {Model A};
  \node[rectangle, draw, fill=Aquamarine!20, right=of M] (MB) {Model B};
  \node[rectangle, draw, fill=Aquamarine!30, below right=of M] (MC) {Model C};

  \draw[->, thick, shorten <=1pt, shorten >=1pt]
    (M.east) to[out=20,in=150]
    node[midway, yshift=1mm, cylinder, shape border rotate=90, draw, fill=blue!30,
         minimum height=6mm,minimum width=4mm, font=\tiny] {A}
    node[midway, below, xshift=8pt,  font=\tiny] {SFT}
    (MA.west);

  \draw[->, thick, shorten <=1pt, shorten >=1pt]
    (M.east) --
    node[midway, yshift=0mm, cylinder, shape border rotate=90, draw, fill=blue!20,
         minimum height=6mm,minimum width=4mm, font=\tiny] {B}
    node[midway, below, xshift=12pt, font=\tiny] {LoRA}
    (MB.west);

  \draw[->, thick, shorten <=1pt, shorten >=1pt]
    (M.east) to[out=-20,in=-150]
    node[midway, yshift=0mm, cylinder, shape border rotate=90, draw, fill=blue!10,
         minimum height=6mm,minimum width=4mm, font=\tiny] {C}
    node[midway, below, xshift=12pt, font=\tiny] {RLHF}
    (MC.west);
}
\vspace{-0.5em}
\\

\midrule
\vspace{1.5em}\textbf{Multi-Agent Orchestration (MAO)}
&
\vspace{2.0em}
\shortstack[c]{
    {\color{purple!50}Agent$_1$} $\Leftrightarrow$ {\color{purple!50}Agent$_2$}\\
    (message exchange)
  }
& \begin{minipage}[t]{\linewidth}
    \vspace{-2em}
    \begin{itemize}[nosep,leftmargin=*]
      \item Multi-Agent Systems
      \item Self-Reflection
      \item Multi-Agent Debate
      \item Chain-of-Thought Ensemble
      \item Function / Tool Calling / MCP
    \end{itemize}
  \end{minipage}
&
\vspace{1em}
\tikz[baseline=(A0.base),node distance=8mm]{%
  \foreach \i/\x in {0/0,1/1.2,2/2.4}{%
    \node[circle,draw,fill=purple!20] (A\i) at (\x cm,0) {\faRobot};
  }
  \draw[<->,thick] (A0) -- (A1);
  \draw[<->,thick] (A1) -- (A2);
} \\

\midrule
\textbf{Multi-Agent Self-Evolving (MASE)}
& \vspace{-0.5em}\shortstack[c]{
    {\color{purple!50}Agents} $\Leftrightarrow$ {\color{teal!50}Environment} \\
    (signals from env.)
  }
&
\vspace*{-\baselineskip}
\begin{minipage}[t]{\linewidth}
\vspace{-3em}
    \begin{itemize}[nosep,leftmargin=*]
      \item Behaviour Optimisation
      \item Prompt Optimisation
      \item Memory Optimisation
      \item Tool Optimisation
      \item Agentic Workflow Optimisation
    \end{itemize}
  \end{minipage}
&
\tikz[baseline=(env.base),node distance=4mm]{%
  \node[cloud, draw, fill=selfevolagent_lighter, inner sep=13pt, minimum width=40mm, minimum height=30mm] (env) {Env.};

  \node[circle, draw, fill=purple!20, minimum size=6mm] (A1) at ($(env.west)+(10mm,6mm)$) {\faRobot};
  \node[circle, draw, fill=purple!20, minimum size=6mm] (A2) at ($(env.east)+(-10mm,6mm)$) {\faRobot};
  \node[circle, draw, fill=purple!20, minimum size=6mm] (A3) at ($(env.north)+(0,-24mm)$) {\faRobot};

  \draw[->, thick, shorten <=1pt, shorten >=1pt] (A1) -- (A2);
  \draw[->, thick, shorten <=1pt, shorten >=1pt] (A2) -- (A3);
  \draw[->, thick, shorten <=1pt, shorten >=1pt] (A3) -- (A1);
}
\vspace{-0.5em}
 \\

\bottomrule
\end{tabular}%
}
\vspace{-5pt}
\caption{Comparison of four LLM‐centric learning paradigms -- Model Offline Pretraining (MOP), Model Online Adaptation (MOA), Multi‐Agent Orchestration (MAO), and Multi‐Agent Self‐Evolving (MASE), highlighting each paradigm’s interaction \& feedback mechanisms, core techniques, and illustrative diagrams to trace the progression from static model training to dynamic, autonomous agent evolution.}
\label{tab:learning-paradigms}
\vspace{-10pt}
\end{table}

Despite self-evolving AI agents representing an ambitious vision for future AI systems, achieving this level of autonomy remains a long-term goal. Current systems are still far from exhibiting the full capabilities required for safe, robust and open-ended self-evolution. 
In practice, current progress towards this vision is achieved through \textit{agent evolution and optimisation} techniques, which provide practical means for enabling agent systems to iteratively refine their components based on interaction data and environmental feedback, thereby enhancing their effectiveness in real-world tasks. Recent research has explored several key directions in this area. One line of work focuses on enhancing the underlying LLM itself to improve the core capabilities, such as planning~\citep{qiao2024agent}, reasoning~\citep{zelikman2022,tong2024dartmath}, and tool use~\citep{feng2025retool}. Another line of research targets the optimisation of auxiliary components within agent systems, including prompts~\citep{xu2022gps,prasad2023grips,yang2024large,wang2025evoagentx}, tools~\citep{yuan-etal-2025-easytool,qu2025from}, memory~\citep{zhong2023memorybankenhancinglargelanguage,lee2024humaninspiredreadingagentgist}, and etc., allowing the agents to better generalise to new tasks and dynamic environments. Furthermore, in multi-agent systems, recent work investigates the optimisation of agent topologies and communication protocols~\citep{bo2024reflective,chen2024optima,zhang2025aflow,zhou2025multi}, aiming to identify agent structures that are best suited to the current task and improve the coordination and information sharing among agents. 

\begin{figure}[t]
    \centering
    \vspace{-5pt}
    \includegraphics[width=\linewidth]{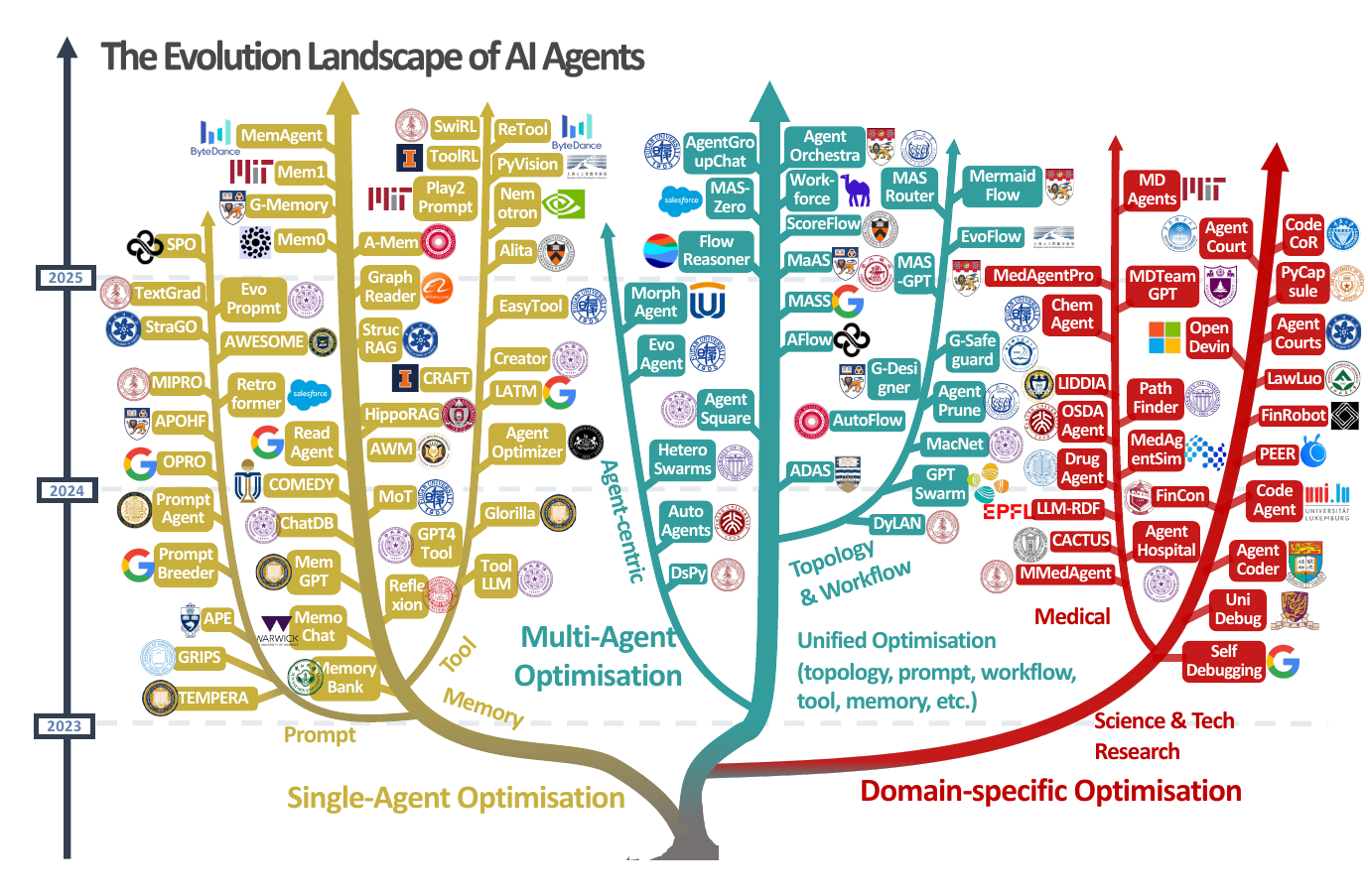}
    \vspace{-20pt}
    \caption{A visual taxonomy of AI agent evolution and optimisation techniques, categorised into three major directions: single-agent optimisation, multi-agent optimisation, and domain-specific optimisation. The tree structure illustrates the development of these approaches from 2023 to 2025, including representative methods within each branch.}
    \label{fig:evolving_tree}
\end{figure}

Existing surveys on AI agents either focus on the general introduction of agent architectures and functionalities~\citep{wang2024survey,guo2024large,xi2025the,luo2025large,liu2025advances,liu2025sew}, or target specific components such as planning~\citep{huang2024understanding}, memory~\citep{zhang2024survey}, collaboration mechanism~\citep{tran2025multi}, and evaluation~\citep{yehudai2025survey}. Other surveys investigate domain-specific applications of agents, such as operating system agents~\citep{hu2025agents} and healthcare agents~\citep{sulis2023survey}.   While these surveys provide valuable insights into various aspects of agent systems, recent advances in agent self-evolution and continual adaptation have not been sufficiently covered, which corresponds to the capabilities of agents that are central to the development of lifelong, autonomous AI systems. This leaves a critical gap in the literature for researchers and practitioners seeking a holistic understanding of the new learning paradigm that underpins adaptive and self-evolving agentic systems. 

To bridge this gap, this survey provides a focused and systematic review of techniques that enable agents to evolve and improve themselves based on interaction data and environmental feedback. Specifically, we introduce a \textit{unified conceptual framework} that abstracts the feedback loop underlying the design of self-evolving agentic systems. This framework identifies four core components: \textit{System Inputs}, \textit{Agent System}, \textit{Environment}, and \textit{Optimisers}, highlighting the evolution loop of agent systems. Building on this framework, we systematically examine a wide range of evolution and optimisation techniques that target different components of the agent systems, including the LLM, prompts, memory, tools, workflow topologies, and communication mechanisms. Moreover, we also investigate domain-specific evolution strategies developed for specialised fields. In addition, we provide a dedicated discussion on the \textit{evaluation}, \textit{safety}, and \textit{ethical considerations} for self-evolving agentic systems, which are critical to ensuring their effectiveness and reliability. 
As a concurrent work, \cite{gao2025survey} surveys self-evolving agents organised around three foundational dimensions: what to evolve, when to evolve, and how to evolve. While their taxonomy offers valuable insights, our survey aims to provide a more comprehensive and integrative perspective, i.e., the unified conceptual framework, on the mechanisms and challenges associated with building lifelong, self-evolving agentic systems. 

This survey aims to provide a comprehensive and systematic review of existing techniques for self-evolving agentic systems, thereby offering researchers and practitioners valuable insights and guidelines for developing more effective and sustainable agentic systems. Figure~\ref{fig:evolving_tree} presents a visual taxonomy of existing agent evolution strategies across single-agent, multi-agent, and domain-specific optimisation, highlighting representative approaches in each direction. 
Our main contributions are as follows: 

\begin{itemize}
    \item We formalise the \ThreeLaws{} and map the evolution of LLM-centric learning paradigms from static pretraining to fully autonomous, lifelong self-evolving agentic systems.
    \item We introduce a unified conceptual framework that abstracts the feedback loop underlying self-evolving agentic systems, and provides a foundation for systematically understanding and comparing different evolution and optimisation approaches. 
    \item We conduct a systematic review of existing evolution and optimisation techniques across single-agent, multi-agent, and domain-specific settings. 
    \item We provide a comprehensive review of evaluation, safety, and ethical considerations for self-evolving agentic systems, emphasising their critical role in ensuring the effectiveness, safety, and responsible deployment of these systems. 
    \item We identify key open challenges and outline promising research directions in agent self-evolution, aiming to facilitate future exploration and advance the development of more adaptive, autonomous, and self-evolving agentic systems. 
\end{itemize}

The remainder of this survey is organised as follows. 
Section~\ref{sec:preliminaries} presents preliminaries on AI agents and multi-agent systems, including their definitions, key components, representative architectures, and the broader vision of autonomous and self-evolving agent systems. 
Section~\ref{sec:conceptual_framework} introduces a unified conceptual framework for agent evolution approaches, outlining the key elements such as system inputs, evolution objectives, agent structures, and optimisers. 
Section~\ref{sec:agent_optimisation} focuses on the optimisation of single-agent systems. It discusses several key aspects such as the optimisation of reasoning strategies, prompt formulation, memory mechanisms, and tool usage. 
Section~\ref{sec:multi_agent_optimisation} focuses on multi-agent systems and review methods for optimising agent workflows, topologies, and inter-agent communication strategies. 
Section~\ref{sec:domain-specific_optimisation} highlights domain-specific agent optimisation techniques and their applications, while Section~\ref{sec:evaluation} discusses evaluation methodologies and benchmarks for assessing agent systems. Section~\ref{sec:challenges_and_directions} presents existing challenges in the agent evolution and optimisation field and outlines some promising future research directions. Finally, we conclude the survey in Section~\ref{sec:conclusions}.

\section{Foundation of AI Agent Systems}
\label{sec:preliminaries} 
To facilitate a clear understanding of agent evolution and optimisation, this section provides an overview of existing AI agent systems. We begin by introducing single-agent systems in Section~\ref{subsec:sas}, outlining their definitions and core components. 
We then turn to multi-agent systems (MAS) in Section~\ref{subsec:mas}, highlighting their motivations, structural paradigms, and collaboration mechanisms. Finally, we present the vision of lifelong, self-evolving agentic systems in Section~\ref{subsec:vision}. 

\subsection{AI Agents}
\label{subsec:sas}
An AI agent refers to an autonomous system capable of perceiving its inputs, reasoning about goals, and interacting with the environment to complete tasks~\citep{luo2025large}. In this section, we focus on single-agent systems, which serve as the foundation of AI agent research. While our goal here is to provide only a brief overview, readers may refer to existing surveys for more comprehensive discussions of AI agent architectures and capabilities~\citep{guo2024large,xi2025the,luo2025large,liu2025advances}. 

An AI agent is typically composed of multiple components that work together to enable autonomous decision-making and execution. The core component of an agent is the \textbf{Foundation Model}, most commonly an \textbf{LLM}\footnote{While this survey focuses on LLMs, the backbone can be any foundation model (e.g., vision–language models, protein sequence/structure models), and the core agentic principles we discuss readily generalise to such backbones.}, which serves as the central reasoning engine responsible for interpreting instructions, generating plans, and producing actionable responses. In addition, there are also some supporting modules that enhance the agent's ability in complex and dynamic environments: 

\begin{enumerate}[label=(\arabic*)]

\item \textbf{Perception Module.} The perception module is responsible for acquiring and interpreting information from the environment~\citep{li2024survey}. This includes processing textual inputs, audio signals, video frames, or other sensory-like data to build a representation suitable for reasoning. 

\item \textbf{Planning Module.} The planning module enables the agent to decompose complex tasks into actionable sub-tasks or sequences of operations and guide their execution across multiple steps~\citep{huang2024understanding}. This process facilitates hierarchical reasoning and ensures coherent task completion. One of the simplest forms of planning involves linear task decomposition, where a problem is broken down into multiple intermediate steps, and the LLM follows these steps to address the problem. This is exemplified by methods such as chain-of-thought prompting~\citep{wei2022chain}. Beyond static planning, more dynamic approaches interleave planning and execution in an iterative loop. For instance, the ReAct~\citep{yao2023react} framework combines reasoning with actions, allowing the agent to revise its plans based on real-time feedback. In addition to linear planning, some methods adopt a branching strategy, where each step may lead to multiple possible continuations. Representative examples are Tree-of-Thought~\citep{yao2023tree} and Graph-of-Thought~\citep{besta2024graph}, which enable the agent to explore multiple reasoning paths. 
    
\item \textbf{Memory Module.} The memory module enables the agent to retain and recall past experience, enabling context-aware reasoning and long-term consistency. Broadly, memory can be categorised into short-term and long-term memory. Short-term memory typically stores the context and interactions generated during the execution of the current task. Once the task is completed, the short-term memory will be removed. 
In contrast, long-term memory persists over time and may store accumulated knowledge, past experiences, or reusable information across tasks. To access relevant long-term memory, many agent systems adopt a retrieval-augmented generation (RAG) module~\citep{zhang2024survey}, where the agent retrieves relevant information from the memory and incorporates them into the input context for the LLM. Designing an effective memory module involves several challenges, including how to structure memory representations, when and what to store, how to retrieve relevant information efficiently, and how to integrate it into the reasoning process~\cite{zeng2024structural}. For a more comprehensive review of memory mechanisms in AI agents, we refer readers to the survey by~\cite{zhang2024survey}. 
    
\item \textbf{Tool Use.} The ability to use external tools is a key factor for AI agents to effectively operate in real-world scenarios. While LLMs are powerful in language understanding and generation, their capabilities are inherently limited by their static knowledge and reasoning capabilities. By using external tools, agents can extend their functional scope, allowing them to better interact with real-world environments. Typical tools include web search engines~\citep{li2025webthinker}, code interpreters or execution environments~\citep{islam2024mapcoder}, and browser automation framework~\citep{browser_use2024}. The design of the tool-use component often involves selecting tools, constructing tool-specific inputs, invoking API calls, and integrating tool outputs back into the reasoning process.

\end{enumerate}

\subsection{Multi-Agent Systems}
\label{subsec:mas}
While single-agent systems have demonstrated strong capabilities in various tasks, many real-world tasks demand specialisation and coordination that exceed the capabilities of a single agent. This limitation has motivated the development of \textit{Multi-Agent Systems} (MAS), which mirror the distributed intelligence found in biological and social systems.

MAS are formally defined as a collection of autonomous agents that interact within a shared environment to achieve goals that are beyond the capabilities of a single agent.
In contrast to single-agent systems that rely solely on individual reasoning and capabilities, MAS focuses on achieving collective intelligence through structured coordination and collaboration among different agents~\citep{tran2025multi}. A fundamental mechanism enabling such coordination is the concept of \textit{agent topology}, the structural configuration that defines how agents are connected and communicate within the system. The topology determines the information flow and collaboration strategies among agents, directly influencing how tasks are distributed and executed. Therefore, MAS is often realised as a \textit{multi-agent workflow}, where the system's topology orchestrates the interactions among agents to accomplish complex, shared goals. 
The key insight is that when multiple agents collaborate through such workflows, the system's overall performance can exceed the sum of the individual capabilities of all agents within the system~\citep{lin2025creativityllmbasedmultiagentsystems,luo2025large}.

MAS brings several notable advantages over single-agent systems. First, MAS can decompose complex tasks into manageable sub-tasks and assign them to specialised agents, which is helpful to improve the overall performance~\citep{krishnan2025advancing,sarkar2025surveyllmagentcommunication}. This approach mirrors human organisational collaboration, enabling MAS to handle tasks that are beyond the capacity of a single agent. Second, MAS supports parallel execution, allowing multiple agents to work simultaneously to complete the task. This feature is particularly advantageous for time-sensitive applications, as it greatly accelerates the problem-solving process~\citep{zhang-etal-2025-parallelized,liu2025advances,li2025parallelizedplanningactingefficientllmbased}. Third, the decentralized nature of MAS enhances robustness: when one agent fails, other agents can dynamically redistribute tasks and compensate for the failure, ensuring graceful degradation rather than a complete system breakdown~\citep{huang2025resiliencellmbasedmultiagentcollaboration,yang2025agentnetdecentralizedevolutionarycoordination}. Fourth, MAS offers inherent scalability, as new agents can be seamlessly integrated without redesigning the entire system~\citep{han2025llmmultiagentsystemschallenges,chen2024internetagentsweavingweb}. Finally, collaborative mechanisms like debate and iterative refinement enable MAS to generate more innovative and reliable solutions by leveraging diverse perspectives and critical evaluation among agents~\citep{guo2024large,lin2025creativityllmbasedmultiagentsystems}. Frameworks such as CAMEL and AutoGen have further streamlined the development of MAS by providing modular architectures, role-playing patterns, and automated orchestration capabilities that reduce engineering overhead~\citep{li2023camel,wu2024autogen}.

\subsubsection{System Architecture}
The architectural design of MAS fundamentally determines how agents organise, coordinate, and execute tasks. These structures range from rigid hierarchies to flexible peer-to-peer networks, each embodying different philosophies about control, autonomy, and collaboration.
\begin{enumerate}[label=(\arabic*)]
    \item \textbf{Hierarchical Structure.} These systems employ static hierarchical organisations, typically linear or tree-based, where tasks are explicitly decomposed and sequentially assigned to specific agents. For instance, MetaGPT~\citep{hong2023metagpt} introduces Standard Operating Procedures (SOPs) to streamline software development, while HALO~\citep{hou2025halohierarchicalautonomouslogicoriented} incorporates Monte Carlo Tree Search to enhance reasoning performance. This highly customised approach offers modularity, ease of development, and domain-specific optimisation, making it prevalent in software development, medicine, scientific research, and social sciences~\citep{zheng2023chatgpt,park2023generativeagentsinteractivesimulacra,qian2024chatdevcommunicativeagentssoftware,li2024agent,cheng2025hawkhierarchicalworkflowframework}. 
    \item \textbf{Centralised Structure.} This architecture follows a manager-follower paradigm where a central agent or higher-level coordinator handles planning, task decomposition, and delegation, while subordinate agents execute assigned subtasks. This design effectively balances global planning with specific task execution~\citep{fourney2024magenticonegeneralistmultiagentsolving,smolagents,camel_workforce_docs}. However, the central node creates performance bottlenecks and introduces single-point-of-failure vulnerabilities that compromise system robustness~\citep{ko2025sevensecuritychallengessolved}.
    
   \item \textbf{Decentralised Structure.} In this architecture, agents collaborate as peers in a distributed network, widely adopted in world simulation applications. The absence of central control prevents single-point failures—damage to any node does not paralyse the entire system, eliminating bottlenecks and enhancing robustness~\citep{lu2024morphagentempoweringagentsselfevolving,yang2025agentnetdecentralizedevolutionarycoordination}. However, this introduces challenges in information synchronisation, data security, and increased collaboration costs~\citep{ko2025sevensecuritychallengessolved}. Recent work explores blockchain technology to address these coordination challenges~\citep{geren2024blockchainlargelanguagemodel,10.5555/3709347.3744042}.
\end{enumerate}

\subsubsection{Communication Mechanisms}
The effectiveness of MAS largely depends on how agents exchange information and coordinate actions. Communication methods in MAS have evolved from simple message passing to sophisticated protocols that balance expressiveness, efficiency, and interoperability.
\begin{enumerate}[label=(\arabic*)]
    \item \textbf{Structured Output.} This approach employs formats like JSON~\citep{li2025graphteamfacilitatinglargelanguage,chen2024internetagentsweavingweb}, XML~\citep{zhang2023appagentmultimodalagentssmartphone,kong2025surveyllmdrivenaiagent}, and executable code~\citep{smolagents} for inter-agent communication. The explicit structure and well-defined parameters ensure high machine readability and interpretability, while standardised formats facilitate cross-platform collaboration~\citep{chen2024internetagentsweavingweb}. These characteristics make structured communication ideal for applications demanding precision and efficiency, such as problem-solving and reasoning tasks. The compact information representation further enhances computational efficiency~\citep{wang2024executablecodeactionselicit}.
    \item \textbf{Natural Language.} Natural language communication preserves rich contextual and semantic details, making it particularly suitable for creative tasks, world simulation, and creative writing scenarios~\citep{liu2025advances}. This expressiveness enables nuanced interactions that capture subtle meanings and intentions. However, it introduces challenges including ambiguity, potential misinterpretation, and reduced execution efficiency compared to structured formats~\citep{guo2024large,yang2025surveyaiagentprotocols,kong2025surveyllmdrivenaiagent}.
    \item \textbf{Standardised Protocols.} Recent advances have introduced specialised protocols designed to standardise MAS communication, creating more inclusive and interoperable agent ecosystems: A2A~\citep{A2A_Protocol_2025} standardises horizontal communication through a structured, peer-to-peer task delegation model, enabling agents to collaborate on complex, long-running tasks while maintaining execution opacity. ANP~\citep{ANP_Protocol_2025} implements secure, open horizontal communication for a decentralised "agent internet" through a hierarchical architecture with built-in Decentralised Identity (DID) and dynamic protocol negotiation. MCP~\citep{MCP_Protocol_2025} standardises vertical communication between individual agents and external tools or data resources through a unified client-server interface. Agora~\citep{Agora_Protocol_2025} functions as a meta-protocol for horizontal communication, enabling agents to dynamically negotiate and evolve their communication methods, seamlessly switching between flexible natural language and efficient structured routines.
\end{enumerate}

\subsection{The Vision of Lifelong, Self-Evolving Agentic Systems}
\label{subsec:vision}
The trajectory from Model Offline Pretraining (MOP) through Model Online Adaptation (MOA) and Multi-Agent Orchestration (MAO) has steadily reduced the degree of manual configuration in LLM-based systems. Yet, even the most advanced multi-agent frameworks today often depend on handcrafted workflows, fixed communication protocols, and human-curated toolchains~\citep{talebirad2023multi,zhao2024expel,luo2025large,tran2025multi}. These static elements constrain adaptability, making it difficult for agents to sustain performance in dynamic, open-ended environments where requirements, resources, and goals evolve over time.

The emerging paradigm of Multi-Agent Self-Evolving (MASE) systems addresses these limitations by closing the loop between deployment and continual improvement. In a MASE system, a population of agents is equipped to autonomously refine their prompts, memory, tool-use strategies, and even their interaction topology -- guided by feedback from the environment and higher-level meta-rewards~\citep{novikov2025alphaevolve,zhang2025darwin}. This continuous optimisation process enables agents not merely to adapt once, but to evolve over their lifetime in response to shifting tasks, domains, and operational constraints.

\textit{Lifelong, self-evolving agentic systems} aim to overcome these constraints by embedding a continuous improvement loop into the core of the architecture. Guided by the \ThreeLaws{} -- Endure (safety adaptation), Excel (performance preservation), and Evolve (autonomous optimisation) -- these systems are designed to:
\begin{enumerate}[label=(\Roman*)]
\item Monitor their own performance and safety profile during operation;
\item Preserve or enhance capabilities through controlled, incremental updates;
\item Autonomously adapt prompts, memory structures, tool-use strategies, and even inter-agent topologies in response to shifting tasks, environments, and resources.
\end{enumerate}

Rather than requiring human designers to handcraft every interaction pattern, a lifelong self-evolving system can generate, evaluate, and refine its own agentic configurations, closing the loop between environment feedback, meta-level reasoning, and structural adaptation. This transforms agents from static executors into continually learning, co-evolving participants in their operational ecosystems.

This vision has far-reaching implications. In scientific discovery, self-evolving agent ecosystems could autonomously generate hypotheses, design experiments, and iterate on research workflows. In software engineering, they could co-evolve development pipelines, integrating new tools as they emerge. In human–AI collaboration, they could learn individual preferences and continually personalise interaction styles. Extending beyond the digital realm, such systems could interface with the physical world through robotics, IoT devices, and cyber–physical infrastructures, perceiving environmental changes, acting upon them, and incorporating real-world feedback into their evolutionary loop. By treating agents as reconfigurable computational entities capable of self-evolving, coordination, and long-term adaptation, MASE offers a pathway toward scalable, sustainable, and trustworthy AI -- AI that is not just trained once, but that \textit{lives}, \textit{learns}, and \textit{lasts}.

\section{A Conceptual Framework of MASE}
\label{sec:conceptual_framework}
To provide a comprehensive overview of self-evolving agentic systems, we propose a high-level conceptual framework that abstracts and summarises the key elements underlying the design and implementation of agent evolution and optimisation methods. This framework provides an abstract yet generalisable view of most existing optimisation approaches, thereby enabling a comprehensive understanding of the field and facilitating comparative analysis across different approaches. 

\begin{figure}[t]
    \centering
    \includegraphics[width=0.9\linewidth]{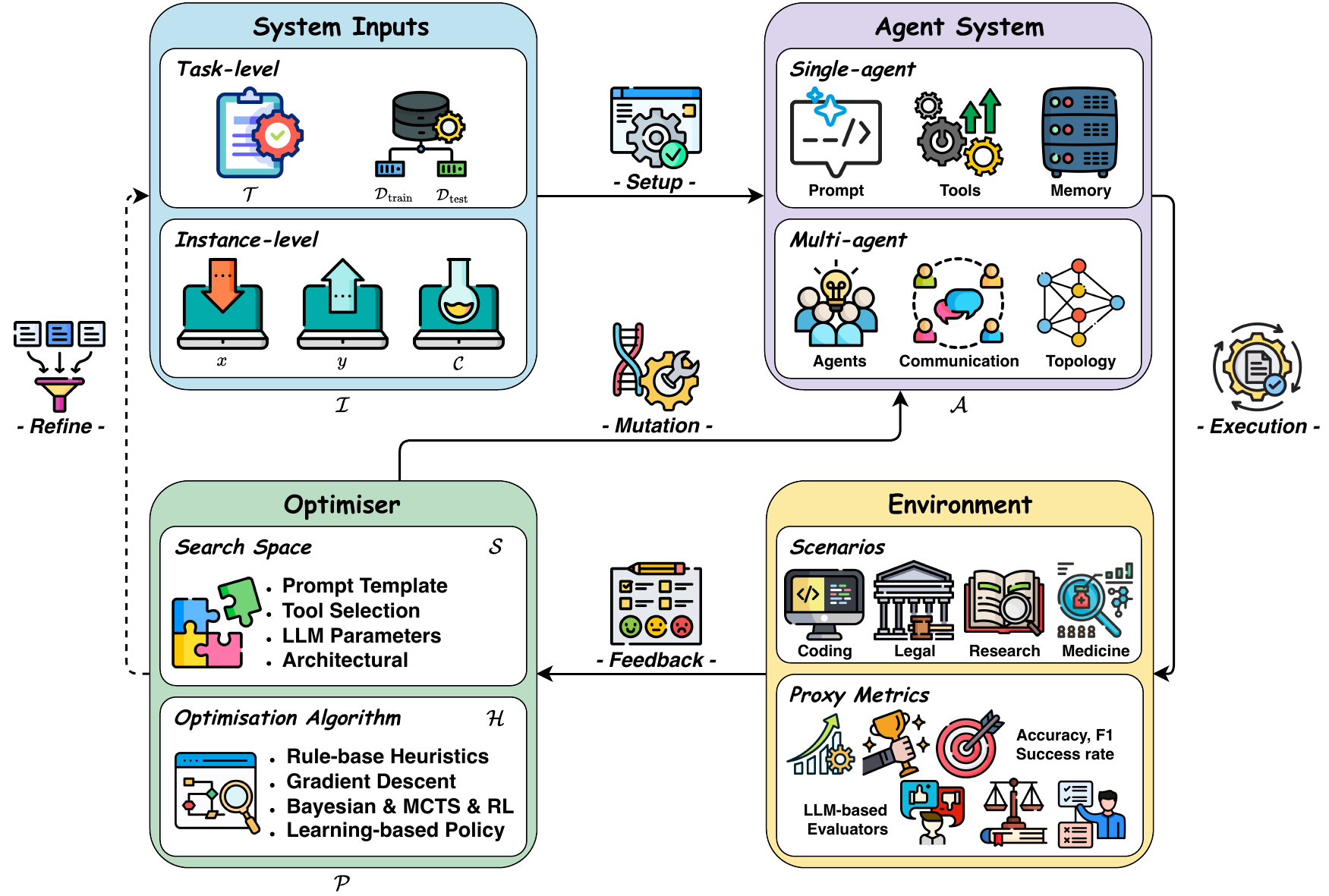}
    \vspace{-3pt}
    \caption{Conceptual framework of the self-evolving process in agent systems. The process forms an iterative optimisation loop comprising four components: \textit{System Inputs, Agent System, Environment}, and \textit{Optimiser}. System inputs define the task setting (e.g., task-level or instance-level). The agent system (in single- or multi-agent form) executes the specified task. The environment (depending on different scenarios) provides feedback via proxy metrics. The optimiser updates the agent system through a defined search space and optimisation algorithm until performance goals are met.}
    \label{fig:Conceptual_Framework}
\end{figure}

\subsection{Overview of the Self-Evolving Process}
We begin with an overview of the self-evolving process in agent systems, which in practice is often realised through iterative optimisation. In this process, the agent system is iteratively updated based on feedback signals obtained from performance evaluations and environmental interactions. 
As illustrated in Figure~\ref{fig:Conceptual_Framework}, the process begins with a task specification, which may include a high-level description, input data, contextual information, or concrete examples. These elements constitute the \textbf{system inputs}, which define the problem setting for the agent system. The \textbf{agent system}, either following a single-agent or multi-agent architecture, is then deployed to perform the task within an \textbf{environment}. The environment provides the operating context and generates feedback signals, which are derived from predefined evaluation metrics, that measure the system's effectiveness and guide subsequent optimisation. Based on feedback from the environment, the \textbf{optimiser} applies specific algorithms and strategies to update the agent system, such as adjusting the LLM parameters, modifying prompts, or refining the system's structure. In some cases, the optimiser may also {refine the system inputs} by synthesising training examples to augment existing datasets, thereby expanding the data available for subsequent optimisation cycle. The updated agent system is then redeployed to the environment, initialising the next iteration. This process forms an iterative, closed feedback loop in which the agent system is progressively refined and optimised over multiple iterations. The loop terminates once a predefined performance threshold is reached or convergence criteria are satisfied.  Building on the conceptual framework of MASE, {\color{selfevolagent_dark!120}\textbf{EvoAgentX}} is the first open-source framework to apply this self-evolving agent process, designed to automate the generation, execution, evaluation, and optimisation of agentic systems~\citep{wang2025evoagentx}.

Building on the overview above, there are four key components within the agent optimisation process: \textit{system inputs}, \textit{agent systems}, \textit{environment} and \textit{optimisers}. In what follows, we provide an introduction to each component, highlighting their individual roles, characteristics and interactions within the optimisation framework. 

\subsection{System Inputs}
System inputs refer to the contextual information and data provided to the optimisation process. 
Formally, we denote the set of system inputs as $\mathcal{I}$, which may consist of one or more elements that specify task requirements, constraints, and available data. These inputs define the problem setting for the agent system and determine the scope of optimisation. 
Depending on the scenario, $\mathcal{I}$ can take different forms: 
\begin{itemize}
    \item \textbf{Task-Level Optimisation}. The most common setting in existing research focuses on improving the agent system's overall performance on a specific task. In this case, the system inputs $\mathcal{I}$ may include a task description $\mathcal{T}$ and a training dataset $\mathcal{D}_{\text{train}}$ used for training or validation: $\mathcal{I} = \{\mathcal{T}, \mathcal{D}_{\text{train}}\}$. A separate test dataset $\mathcal{D}_{\text{test}}$ can also be incorporated to evaluate the optimised agent's performance. In some scenarios, task-specific labeled data, i.e., $\mathcal{D}_{\text{train}}$, are unavailable. To enable optimisation in such settings, recent approaches~\citep{huang2025rzero,zhao2025absolute,liu2025spiral} propose to dynamically \textit{synthesise} training examples, often through LLM-based data generation, to create a surrogate dataset for iterative improvement. 
    \item \textbf{Instance-Level Optimisation}. Recent studies also explore a more fine-grained setting, where the objective is to enhance the agent system's performance on a specific example~\citep{sun2024query,novikov2025alphaevolve}. In this case, the system inputs may consist of an input-output pair $(x, y)$, along with optional contextual information $\mathcal{C}$, i.e., $\mathcal{I} = \{x, y, \mathcal{C}\}$. 
\end{itemize}

\subsection{Agent Systems} 
The agent system is the core component within the feedback loop that is subject to optimisation. It defines the decision-making process and functionality of the agent(s) in response to given inputs. Formally, we denote the agent system as $\mathcal{A}$, which may consist of a single agent or a collection of collaborating agents. The agent system $\mathcal{A}$ can be further decomposed into several components, such as the underlying LLM, prompting strategy, memory module, tool-use policy, etc. Optimisation methods may focus on one or more of these components depending on the intended scope. In most existing works, optimisation is performed on a single component of $\mathcal{A}$, such as finetuning the LLM to enhance reasoning and planning capabilities~\citep{zelikman2022,tong2024dartmath,lai2024stepdpostepwisepreferenceoptimization}, or tuning the prompts and selecting appropriate tools to improve task-specific performance without modifying the LLM itself~\citep{yang2024large,yuan-etal-2025-easytool}. Moreover, recent research has also explored joint optimisation of multiple components with $\mathcal{A}$. For example, in single-agent systems, some approaches jointly optimise LLM and prompting strategy to better align model behaviour with task requirements~\citep{dilara2024fine}. In multi-agent systems, existing studies have explored the joint optimisation of prompts and inter-agent topology to improve overall effectiveness~\citep{zhang2025aflow,zhou2025multi}.

\subsection{Environments} 
The environment serves as the external context in which the agent system operates and generates outputs. Specifically, the agent system interacts with the environment by perceiving its inputs, executing actions, and receiving corresponding outcomes. Depending on the task, the environment can vary from a benchmark dataset to a fully dynamic, real-world setting~\citep{liu2023agentbench}. For example, in code generation tasks, the environment may include code execution and verification components such as compilers, interpreters, and test cases. In scientific research, it may consist of literature databases, simulation platforms, or laboratory equipment. 

Beyond providing the operational context, the environment also plays a critical role in generating $\textit{feedback signals}$ that inform and guide the optimisation process. These signals are typically derived from $\textit{evaluation metrics}$ that quantify the effectiveness or efficiency of the agent system. In most cases, such metrics are task-specific, e.g., accuracy, F1, or success rate, which provide quantitative measures of performance. However, in settings where labelled data or ground-truth are unavailable, LLM-based evaluators are often employed to estimate performance~\citep{yehudai2025survey}. These evaluators can generate proxy metrics or provide textual feedback by assessing aspects such as correctness, relevance, coherence, or alignment with task instructions. A more detailed discussion of evaluation strategies across different applications is presented in Section~\ref{sec:evaluation}.

\subsection{Optimisers}

Optimisers ($\mathcal{P}$) are the core component of the self-evolving feedback loop, responsible for refining the agent system $\mathcal{A}$ based on performance feedback from the environment. Their objective is to search, via specialised algorithms and strategies, for the agent configuration that achieves the best performance under the given evaluation metric. Formally, this can be expressed as:
\begin{align}
\mathcal{A}^* = \arg \max_{\mathcal{A} \in \mathcal{S}} \mathcal{O}(\mathcal{A}; \mathcal{I}),
\end{align}
where $\mathcal{S}$ denotes the search space of configurations, $\mathcal{O}(\mathcal{A}; \mathcal{I}) \in \mathbb{R}$ is the evaluation function that maps the performance of $\mathcal{A}$ on the given system inputs $\mathcal{I}$ to a scalar score, and $\mathcal{A}^*$ denotes the optimal agent configuration. 

An optimiser is typically defined by two core components: 
(1) \textbf{search space ($\mathcal{S}$)}: This defines the set of agent configurations that can be explored and optimised. The granularity of $\mathcal{S}$ depends on which part(s) of the agent system are subject to optimisation, ranging from agent prompts or tool selection strategies to continuous LLM parameters or architectural structures. 
(2) \textbf{optimisation algorithm ($\mathcal{H}$)}: This specifies the strategy used to explore $\mathcal{S}$ and select or generate candidate configurations. It can include rule-based heuristics, gradient descent, Bayesian optimisation, Monte Carlo Tree Search (MCTS), reinforcement learning, evolutionary strategies, or learning-based policies. Together, the pair $(\mathcal{S}, \mathcal{H})$ defines the behaviour of the optimiser and determines how efficiently and effectively it can adapt the agent system toward better performance. 

\begin{figure}[!t]
    \centering
    \resizebox{0.97\textwidth}{!}{
    \begin{forest}
        for tree={
            forked edges,
            grow'=0,
            draw,
            rounded corners,
            node options={align=center},
            font=\Large,
            minimum height=6mm,
            text width=1.5cm,
            s sep=5pt,
            l sep=10pt, 
            calign=child edge,
            calign child=(n_children()+1)/2,
        },
        [Agentic \\Self-Evolution, root
            [Single-Agent Optimisation, data_section
                [Behaviour \\Optimisation, data_parent_section
                    [SFT, data_section 
                        [STaR \citep{zelikman2022}; ToRA~\citep{tora}; NExT~\citep{next};
                        ,data_section, text width=16.5cm
                        ]
                    ]
                    [RL, data_section  
                        [Self-Rewarding~\citep{yuan2024self}; DeepSeek-Prover-v1.5~\citep{xin2025deepseekproverv};
                        Absolute-Zero~\citep{zhao2025absolute}
                        ,data_section, text width=16.5cm
                        ]
                    ]
                    [Verifier Module, data_section
                        [Baldur~\citep{first2023baldur}; Math-Shepherd~\citep{wang-etal-2024-math}; Rewarding Progress~\citep{setlur2025rewarding};
                        ,data_section, text width=16.5cm
                        ]
                    ]
                    [Search-Based, data_section
                        [CoT-SC~\citep{wang2023self}; Tree-of-Thoughts~\citep{yao2023tree}; Graph-of-Thoughts~\citep{besta2024graph};
                        ,data_section, text width=16.5cm
                        ]
                    ]
                ]
                [Prompt \\Optimisation, data_parent_section
                    [Edit-Based, data_section
                        [GPS~\citep{xu2022gps}; GrIPS~\citep{prasad2023grips}; TEMPERA~\citep{zhang2023tempera};
                        ,data_section, text width=16.5cm
                        ]
                    ]
                    [Generation-Based, data_section
                        [APE~\citep{zhou2023large}; PromptAgent~\citep{wang2024promptagent}; OPRO~\citep{yang2024large}; APOHF~\citep{lin2024prompt};\\
                         RETROFORMER~\citep{yao2024retroformer}; MIPRO~\citep{opsahl2024optimizing}; StraGo~\citep{wu2024strago}; SPO~\citep{xiang2025self};
                        ,data_section, text width=16.5cm
                        ]
                    ]
                    [Text Gradient-Based, data_section
                        [ProTeGi~\citep{pryzant2023automatic}; TextGrad~\citep{yuksekgonul2024textgrad};
                        ,data_section, text width=16.5cm
                        ]
                    ]
                    [Evolution-Based, data_section
                        [EvoPrompt~\citep{guo2024evoprompt}; Promptbreeder~\citep{fernando2024promptbreeder};
                        ,data_section, text width=16.5cm
                        ]
                    ]
                ]
                [Memory \\Optimisation, data_parent_section
                    [Short-term Memory, data_section
                        [COMEDY~\citep{chen2024compressimpressunleashingpotential};ReadAgent~\citep{lee2024humaninspiredreadingagentgist}; MoT~\citep{li2023motmemoryofthoughtenableschatgpt};StructRAG~\citep{li2024structragboostingknowledgeintensive};MemoryBank~\citep{zhong2023memorybankenhancinglargelanguage};
                        ,data_section, text width=16.5cm
                        ]
                    ]
                    [Long-term Memory, data_section
                        [EWE~\citep{chen2025improvingfactualityexplicitworking}; A-MEM~\citep{xu2025amemagenticmemoryllm}; Mem0~\citep{chhikara2025mem0buildingproductionreadyai}; GraphReader~\citep{li2024graphreaderbuildinggraphbasedagent}; AWM~\citep{wang2024agentworkflowmemory}; MyAgent~\citep{Hou_2024}; 
                        ,data_section, text width=16.5cm
                        ]
                    ]
                ]
                [Tool Optimisation, data_parent_section
                    [Training-Based, data_section
                        [ToolLLM~\citep{qin2024toolllm}; Confucius~\citep{gao2024confucius}; ReTool~\citep{feng2025retool}; ToolRL~\citep{qian2025toolrl}; SWiRL~\citep{goldie2025synthetic}; Nemotron-Research-Tool-N1~\citep{zhang2025nemotron};
                        ,data_section, text width=16.5cm
                        ]
                    ]
                    [Prompt-Based, data_section
                        [EASYTOOL~\citep{yuan-etal-2025-easytool}; PLAY2PROMPT~\citep{fang2025play2prompt}; DRAFT~\citep{qu2025from}; JointOptim~\citep{wu2025joint}; ,data_section, text width=16.5cm
                        ]
                    ]
                    [Tool Creation, data_section
                        [CREATOR~\citep{qian2023creator}; LATM~\citep{cai2024large}; CRAFT~\citep{yuan2024craft}; AgentOptimizer~\citep{zhang2024offline}; Alita~\citep{qiu2025alita};
                        ,data_section, text width=16.5cm
                        ]
                    ]
                ]
            ]
            [Multi-Agent Optimisation, model_parent_section
                [Prompt Optimisation, model_parent_section 
                [AutoAgents~\citep{chen2024autoagentsframeworkautomaticagent}; DSPy~\citep{DBLP:journals/corr/abs-2312-13382}; MIPRO~\citep{opsahl2024optimizing}; PromptWizard~\citep{agarwal2024promptwizard},
                model_section, text width=21.1cm] 
                ]
                [Topology\\ Optimisation, model_parent_section
                    [Code-level Workflow, model_section
                        [AutoFlow~\citep{li2024autoflowautomatedworkflowgeneration}; AFlow~\citep{zhang2025aflow}; ScoreFlow~\citep{wang2025scoreflow}; MAS\mbox{-}GPT~\citep{ye2025masgpttrainingllmsbuild};
                        ,model_section, text width=16.5cm
                        ]
                    ]
                    [Communication Graph, model_section
                        [GPTSwarm~\citep{zhuge2024gptswarm}; DynaSwarm~\citep{leong2025dynaswarmdynamicallygraphstructure}; G\mbox{-}Designer~\citep{zhang2025gdesignerarchitectingmultiagentcommunication}; NetSafe~\citep{yu2024netsafeexploringtopologicalsafety}; AgentPrune~\citep{zhang2025cut}; AGP~\citep{li2025adaptivegraphpruningmultiagent};
                        ,model_section, text width=16.5cm
                        ]
                    ]
                ]
                [Unified \\Optimisation, model_parent_section
                    [Code-Based, model_section
                        [ADAS~\citep{hu2025automated}; FlowReasoner~\citep{zhang2025evoflowevolvingdiverseagentic}; 
                        ,model_section, text width=16.5cm
                        ]
                    ]
                    [Search-Based, model_section
                        [EvoAgent~\citep{yuan-etal-2025-evoagent}; MASS~\citep{zhou2025multi}; DebFlow~\citep{su2025debflowautomatingagentcreation}; EvoFlow~\citep{zhang2025evoflowevolvingdiverseagentic}; MAS-ZERO \citep{ke2025maszerodesigningmultiagentsystems};
                        ,model_section, text width=16.5cm
                        ]
                    ]
                    [Learning-Based, model_section
                        [MaAS~\citep{zhang2025multiagentarchitecturesearchagentic}; ANN~\citep{ma2025agenticneuralnetworksselfevolving}; 
                        ,model_section, text width=16.5cm
                        ]
                    ]
                ]
                [LLM Backbone Optimisation, model_parent_section
                    [Reasoning-Oriented, model_section
                        [AutoFlow~\citep{li2024autoflowautomatedworkflowgeneration}; AFlow~\citep{zhang2025aflow}; ScoreFlow~\citep{wang2025scoreflow}; MAS\mbox{-}GPT~\citep{ye2025masgpttrainingllmsbuild};
                        ,model_section, text width=16.5cm
                        ]
                    ]
                    [Collaboration-Oriented, model_section
                        [COPPER~\citep{bo2024reflective}; OPTIMA~\citep{chen2024optima}; MaPoRL~\citep{park2025maporl}; 
                        ,model_section, text width=16.5cm
                        ]
                    ]
                ]
            ]
            [Domain-Specific Optimisation, training_section
                [Biomedicine, training_section
                    [Medical Diagnosis, training_section
                        [MedAgentSim~\citep{almansoori2025self}; PathFinder~\citep{ghezloo2025pathfinder}; MDAgents~\citep{kim2024mdagents}; MDTeamGPT~\citep{chen2025mdteamgpt}; MMedAgent~\citep{li2024mmedagent}; MedAgent-Pro~\citep{wang2025medagent}; ,training_section, text width=16.5cm
                        ]
                    ]
                    [Molecular Discovery, training_section
                        [CACTUS~\citep{mcnaughton2024cactus}; LLM-RDF~\citep{m2024augmenting}; ChemAgent~\citep{tang2025chemagent}; OSDA Agent~\citep{hu2025osda}; DrugAgent~\citep{inoue2025drugagent};LIDDIA~\citep{averly2025liddia};, training_section, text width = 16.5cm]
                    ]
                ]
                [Programming, training_section
                    [Code Refinement, training_section
                        [Self-Refine~\citep{madaan2023self}; AgentCoder~\citep{huang2023agentcoder}; CodeAgent~\citep{tang2024codeagent}; CodeCoR~\citep{pan2025codecor}; OpenHands~\citep{wang2024openhands};
                            ,training_section, text width=16.5cm
                        ]
                    ]
                    [Code Debugging, training_section
                        [Self-Debugging~\citep{chen2023teaching}; Self-Edit~\citep{zhang2023self}; PyCapsule~\citep{adnan2025large}; RGD~\citep{jin2024rgd};  , training_section, text width=16.5cm
                        ]
                    ]
                ]
                [Financial and Legal Research, training_section
                    [Financial Decision-Making, training_section
                    [FinCon~\citep{yu2024fincon}; PEER~\citep{wang2024peer}; FinRobot~\citep{yang2024finrobot}; , training_section, text width=16.5cm
                        ]
                    ]
                    [Legal Reasoning, training_section
                    [LawLuo~\citep{sun2024lawluo};AgentCourt~\citep{chen2024agentcourt};LegalGPT~\citep{shi2024legalgpt}; ,
                        training_section, text width=16.5cm
                        ]
                    ]
                ]          
            ]
        ]
    \end{forest}
    }
    \caption{A comprehensive hierarchical categorisation of Agentic Self-Evolution methods, encompassing single-agent, multi-agent and domain-specific optimisation categories, illustrated with selected representative works.}
    \label{fig:agent-evaluation-typology}
\end{figure}
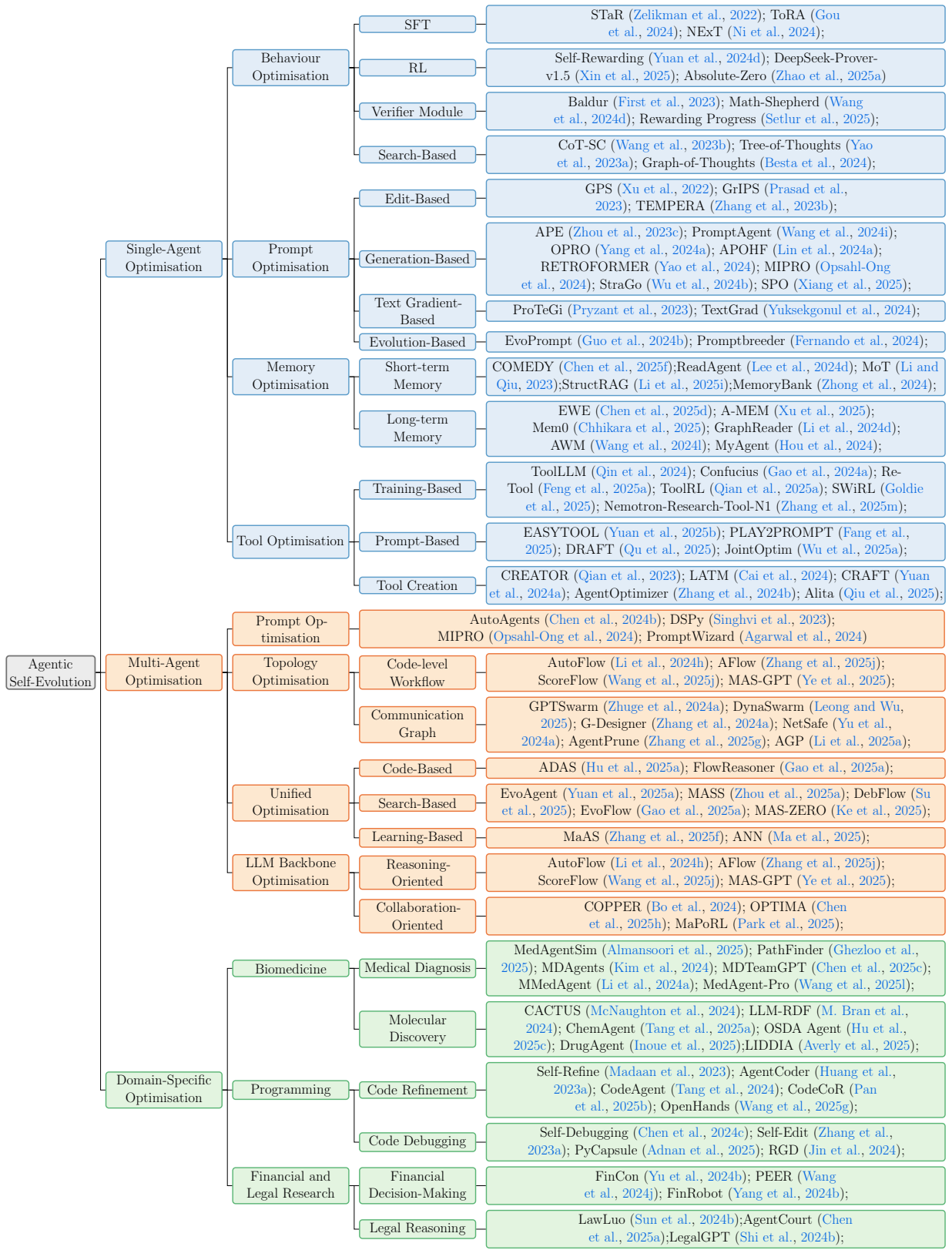

In the following sections, we introduce typical optimisers in three different settings: single-agent systems (Section~\ref{sec:agent_optimisation}), multi-agent systems (Section~\ref{sec:multi_agent_optimisation}), and domain-specific agent systems (Section~\ref{sec:domain-specific_optimisation}). Each setting exhibits distinct characteristics and challenges, leading to different designs and implementations of optimisers. In single-agent optimisation, the focus is on improving an individual agent's performance by tuning LLM parameters, prompts, memory mechanisms, or tool-use policies. In contrast, multi-agent optimisation extends the scope to optimising not only individual agents but also their structural designs, communication protocols, and collaboration capabilities. Domain-specific agent optimisation presents additional challenges, where optimisers must account for specialised requirements and constraints inherent to particular domains, leading to tailored optimiser designs. A comprehensive hierarchical categorisation of these optimisation settings and representative methods is provided in Figure~\ref{fig:agent-evaluation-typology}.

\section{Single-Agent Optimisation}
\label{sec:agent_optimisation} 

\begin{figure}[H]
    \centering
    \vspace{-1pt}
    \includegraphics[width=0.95\linewidth]{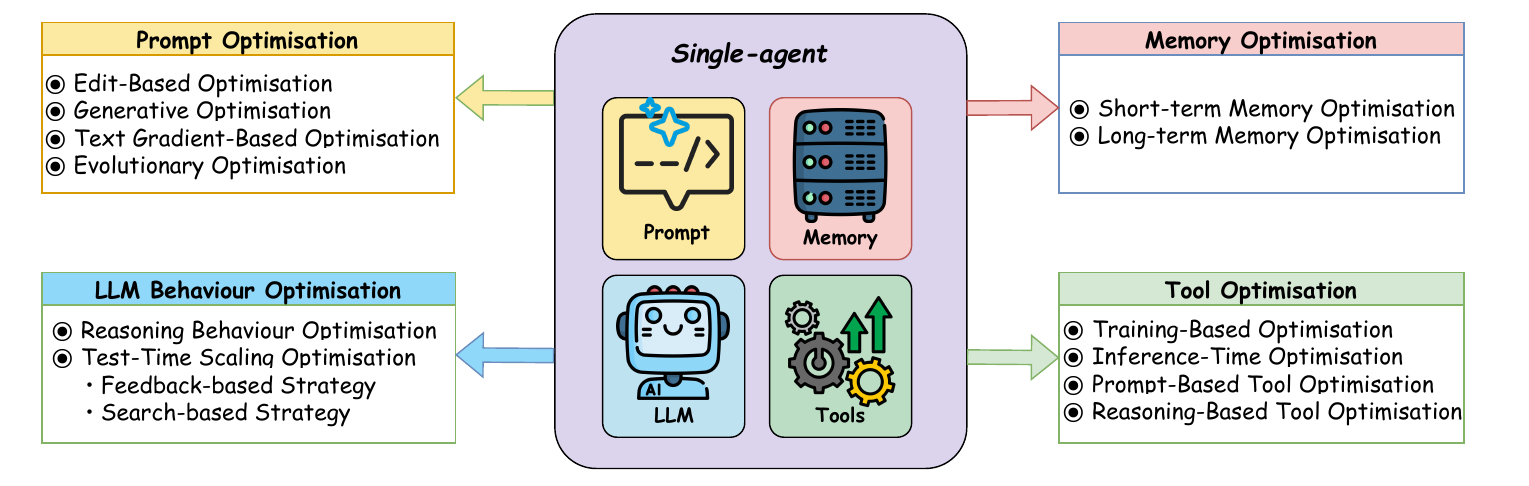}
    \caption{An overview of single-agent optimisation approaches, categorised by the targeted component within the agent system: prompt, memory, and tool.}
    \label{fig:Single_Agent_Optimisation}
    \vspace{-2pt}
\end{figure}

Single-agent optimisation focuses on improving the performance of a single-agent system. According to the optimisation feedback loop introduced earlier, the key challenge lies in the design of optimisers for updating the system. This involves identifying the specific components of the agent system to optimise (i.e., search space), determining the particular capabilities to enhance, and choosing appropriate optimisation strategies to effectively achieve these improvements (i.e., optimisation algorithm). 

In this section, we organise single-agent optimisation approaches based on the targeted component within the agent system, as this determines both the structure of the search space and the choice of optimisation methods. Specifically, we focus on four major categories: 
(1) \textit{LLM Behaviour optimisation}, which aims to improve the LLM's reasoning and planning capabilities through either parameter tuning or test-time scaling techniques; 
(2) \textit{Prompt optimisation}, which focuses on adapting prompts to guide the LLM towards producing more accurate and task-relevant outputs; 
(3) \textit{Memory optimisation}, which aims to enhance the agent's ability to store, retrieve, and reason over historical information or external knowledge; 
(4) \textit{Tool optimisation}, which focuses on enhancing the agent's ability to effectively leverage existing tools, or autonomously create or configure new tools to accomplish complex tasks. 
Figure \ref{fig:Single_Agent_Optimisation} shows the major categories of single-agent optimisation approaches.

\definecolor{paired-light-blue}{RGB}{198, 219, 239}
\definecolor{paired-dark-blue}{RGB}{49, 130, 188}
\definecolor{paired-light-orange}{RGB}{251, 208, 162}
\definecolor{paired-dark-orange}{RGB}{230, 85, 12}
\definecolor{paired-light-green}{RGB}{199, 233, 193}
\definecolor{paired-dark-green}{RGB}{49, 163, 83}
\definecolor{paired-light-purple}{RGB}{218, 218, 235}
\definecolor{paired-dark-purple}{RGB}{117, 107, 176}
\definecolor{paired-light-gray}{RGB}{217, 217, 217}
\definecolor{paired-dark-gray}{RGB}{99, 99, 99}
\definecolor{paired-light-pink}{RGB}{222, 158, 214}
\definecolor{paired-dark-pink}{RGB}{123, 65, 115}
\definecolor{paired-light-red}{RGB}{231, 150, 156}
\definecolor{paired-dark-red}{RGB}{131, 60, 56}
\definecolor{paired-light-yellow}{RGB}{231, 204, 149}
\definecolor{paired-dark-yellow}{RGB}{141, 109, 49}
\definecolor{light-green}{RGB}{118, 207, 180}
\definecolor{raspberry}{RGB}{228, 24, 99}

\tikzset{%
    root/.style =          {align=center,text width=3cm,rounded corners=3pt, line width=0.5mm, fill=paired-light-gray!50,draw=paired-dark-gray!90},
    data_section/.style =  {align=center,text width=4cm,rounded corners=3pt, fill=paired-light-blue!50,draw=paired-dark-blue!80,line width=0.4mm},
    data_parent_section/.style =  {align=center,text width=4cm,font=\Large,minimum height=12mm,rounded corners=3pt, fill=paired-light-blue!50,draw=paired-dark-blue!80,line width=0.4mm},
    model_section/.style = {align=center,text width=4cm,rounded corners=3pt, fill=paired-light-orange!50,draw=paired-dark-orange!80,line width=0.4mm},
    model_parent_section/.style = {align=center,text width=4cm,font=\Large,minimum height=12mm,rounded corners=3pt, fill=paired-light-orange!50,draw=paired-dark-orange!80,line width=0.4mm},
    training_section/.style = {align=center,text width=4cm,rounded corners=3pt, fill=paired-light-green!50,draw=paired-dark-green!80, line width=0.4mm},
    inference_section/.style = {align=center,text width=4cm,rounded corners=3pt, fill=paired-light-red!35,draw=paired-light-red!90, line width=0.4mm},
    discussion_section/.style = {align=center,text width=4cm,rounded corners=3pt, fill=paired-light-purple!35,draw=paired-dark-purple!90, line width=0.4mm},
    subsection/.style =    {align=center,text width=3.5cm,rounded corners=3pt}, 
}

\subsection{LLM Behaviour Optimisation}

Backbone LLMs lay the foundation for single-agent systems, serving as the primary component responsible for planning, reasoning, and task execution. Therefore, enhancing the planning and reasoning capabilities of the LLM is central to improving the overall effectiveness of the agent system. Recent efforts in this direction broadly fall into two categories: (1) \textbf{training-based methods}, which directly update the model's parameters to improve reasoning ability and task performance; (2) \textbf{test-time methods}, which aim to improve LLM's behaviour during inference without modifying its parameters. In the following, we review and summarise representative approaches from both categories. 

\subsubsection{Training-Based Behaviour Optimisation}

While LLMs have demonstrated strong linguistic capabilities~\citep{zhao2023survey}, recent research~\citep{wu-etal-2024-reasoning} highlights a notable gap between their fluency in natural language and their ability to perform complex reasoning. This discrepancy limits the effectiveness of LLM-based agents in tasks that require multi-step inference and complex decision-making. To address this, recent work has explored reasoning-oriented training methods, using supervised fine-tuning (SFT) and reinforcement learning (RL) to help models systematically evaluate and refine their responses.

\paragraph{{Supervised Fine-tuning.}}
The core idea of supervised fine-tuning is to train agents using annotated data that contains detailed reasoning steps, allowing the model to learn a complete mapping from the input question, through intermediate reasoning processes, to the final answer. This approach typically relies on carefully constructed reasoning trajectories, which can typically be constructed from (1) \textbf{rollouts generated by the agent itself during execution}, and (2) \textbf{demonstrations produced by stronger teacher agents}. By imitating these trajectories, the agent acquires the ability to perform step-by-step reasoning in a structured manner. 
STaR~\citep{zelikman2022} proposes an iterative fine-tuning procedure, where the model is trained on instances it has solved correctly and refines incorrect traces to generate better trajectories. Based on this idea, NExT~\citep{next} uses self-generated trajectories filtered by unit test correctness to self-evolve agents for program repair tasks. Similarly, Deepseek-Prover~\citep{xin2024deepseek} progressively evolve the agent by iteratively training the policy model with verified proofs, enabling it to generate increasingly accurate formal proofs for theorem-proving tasks. Another line of work fine-tunes agents on trajectories generated by proprietary LLMs, across domains such as mathematics~\citep{tora, yin2024mumath} and science \citep{ma-etal-2024-sciagent}. Beyond agentic capability, \cite{min2024imitate, huang2024o1replicationjourney,bespoke_stratos} train models based on trajectories generated by OpenAI o1 \citep{openai2024openaio1card} to replicate its thinking capability, aiming to further improve the agent backbone's reasoning ability. 

\paragraph{Reinforcement Learning.}
RL treats reasoning as a sequential decision-making process where the model is rewarded for producing correct or high-quality reasoning paths. 
One of the strategies is preference-based optimisation, where DPO \citep{rafailov2023direct} is applied using preference pairs generated from various sources, such as test case performance, correctness of final outcomes, or pseudo-labels from trained process reward models (PRMs) \citep{hui2024qwen2, min2024imitate, jiao-etal-2024-learning,liu2025learn}. \cite{yuan2024self} further introduce a self-evolving framework where the policy model uses its own judgments to iteratively refine its reasoning ability.
Similarly, Agent Q~\citep{putta2024agent} combines MCTS-guided search and a self-critique mechanism to iteratively improve agents' decision making in web environments via DPO, leveraging both successful and failed trajectories.
In another line of work, Tülu 3~\citep{lambert2024tulu} applies reinforcement learning with verifiable rewards across mathematical and instruction-following tasks without any learned reward model. Notably, DeepSeek-R1~\citep{guo2025deepseek} further demonstrates the feasibility of pure RL with Group Relative Policy Optimisation ~\citep{shao2024deepseekmath} when ground-truth verification is possible. 
Building on this direction, \cite{xin2025deepseekproverv} extend the idea to enhance DeepSeek-Prover by incorporating reinforcement learning from proof assistant feedback. \cite{liu2025diving} further explore self-evolving training in the multimodal setting by introducing MSTAR, a framework that leverages RL to overcome performance saturation and enhance reasoning capabilities through iterative self-improvement. 
Beyond using verifiable rewards in a fixed dataset, Absolute Zero~\citep{zhao2025absolute} trains a single model that alternates between task proposer and solver roles, self-evolving by generating and solving its own problems. 
Similarly, R-Zero~\citep{huang2025rzero} employs a dual-mode framework in which a challenger generates tasks tailored to the solver's current competence, enabling both to evolve iteratively without external supervision. 

\subsubsection{Test-Time Behaviour Optimisation} 

As training resources become increasingly constrained and API-based models cannot be fine-tuned, test-time compute emerges as a solution to these limitations by enabling models to refine or extend their reasoning capabilities during inference \textit{without additional training}. By increasing the inference budget, models are able to ``think longer''. 

Scaling test-time capabilities can be achieved through two primary strategies. The first involves \textit{guiding inference through the incorporation of external feedback}, which facilitates the model's refinement of its responses. The second strategy focuses on \textit{generating multiple candidate outputs using more efficient sampling algorithms}. This is followed by a selection process where a verifier identifies the most suitable output. Notably, these two approaches are in fact closely related. The feedback used to guide generation in the former can naturally serve as a verifier in the latter.

\paragraph{{Feedback-based Strategy.}} A natural method is to adjust a model's behaviour based on the quality of its generated outputs. This process typically relies on feedback from a \textit{verifier}, which provides either an exact or estimated score to guide the model. We categorise feedback into two types. \textit{Outcome-level feedback} provides a single score or signal based on the final output, regardless of the number of reasoning steps taken. For tasks where ground-truth is easily accessible, the verifier can be implemented as an external tool to provide accurate feedback. For example, CodeT \citep{ChenZNZLLC23} and LEVER \citep{ni2023lever} leverage a compiler to execute the generated code and validate its correctness against test cases. START~\citep{li2025start} and CoRT~\citep{li2025cort} employ hint-based tool invocation to enhance long CoT reasoning. Similarly, Baldur \citep{first2023baldur} leverages error messages from proof assistants to further repair incorrect proofs generated by LLMs. However, for most tasks, ground-truth is not always available at inference time. As a result, a more general approach is to train a model to serve as the verifier that assigns a score to each candidate response \citep{liu2024skyworkrewardbagtricksreward, liu2025skyworkrewardv2scalingpreferencedata}, allowing them to be ranked based on predicted quality. However, this form of feedback is relatively sparse, as it evaluates only the final output. In contrast, \textit{step-level feedback} evaluates each intermediate step during the generation process, offering finer-grained supervision. Relying solely on outcome feedback often leads to the \textit{unfaithful reasoning} problem \citep{turpin2023language}, where incorrect reasoning chains may still result in correct final answers. To address this, recent work \citep{wang-etal-2024-math, jiao-etal-2024-learning, setlur2025rewarding} increasingly focuses on training process reward models to detect and correct errors throughout the reasoning process, generally yielding better improvement than using outcome-level feedback.

\paragraph{{Search-based Strategy.}} Complex reasoning tasks often admit multiple valid paths leading to the correct solution. Search-based approaches take advantage of this property by exploring several candidate reasoning trajectories in parallel, enabling the model to better navigate the solution space. With the help of critic models, various search strategies have been developed to guide the decoding process. For example, CoT-SC~\citep{wang2023self} adopts a best-of-N approach: it generates multiple reasoning paths and selects the final answer based on the majority vote over outcomes. DBS \citep{zhu2024deductive} proposes the use of beam search in combination with step-level feedback to refine intermediate reasoning steps, while CoRe \citep{zhu-etal-2023-solving} and Tree-of-Thoughts~\citep{yao2023tree} explicitly model the reasoning process as a tree structure, using Monte Carlo Tree Search (MCST) for a balance between exploration and exploitation during searching.
Forest-of-Thought \citep{bi2025forestofthought} further generalises this idea by enabling multiple trees to make independent decisions and applying a sparse activation mechanism to filter and select outputs from the most relevant trees. Beyond tree-based methods, other approaches have also explored alternative structural formulations of reasoning. Graph-of-Thoughts~\citep{besta2024graph} organises intermediate thoughts as nodes in a graph and applies graph-based operations to support flexible reasoning and information flow. Buffer-of-Thoughts \citep{yang2024buffer} introduces a dynamic memory buffer to store and instantiate meta-level thoughts during reasoning.

\subsection{Prompt Optimisation}
\label{subsec:prompt_optimisation}
In single-agent systems, prompts play a critical role in defining the agent's goals, behaviour, and task-specific strategies. They typically contain instructions, illustrative demonstrations, and contextual information that guide the underlying LLM in generating appropriate outputs. However, it is well-known that LLMs are highly sensitive to prompts; even minor variations in phrasing, formatting, or word ordering can lead to significant changes in the LLMs' behaviour and output~\citep{loya2023exploring, zhou2024batch}. This sensitivity makes it difficult to design robust and generalisable AI agent systems, motivating the development of prompt optimisation techniques to automatically search for high-quality prompts. Prompt optimisation methods can be categorised based on the strategies used to navigate the prompt space and identify high-quality prompts that enhance model performance. In this section, we review and summarise four representative categories: edit-based methods, generative methods, text gradient-based methods, and evolutionary methods. 

\subsubsection{Edit-Based Prompt Optimisation} 
Earlier attempts in prompt optimisation focus on \textit{edit-based} approaches, which iteratively refine human-written prompts through predefined editing operations, such as token insertion, deletion or substitution~\citep{prasad2023grips,pan2024plum,lu2024strings,zhang2023tempera,zhou2023survival,agarwal2024promptwizard}. These methods treat prompt optimisation as a local search problem over prompt space, aiming to gradually improve prompt quality while preserving the core semantics of the original instruction. For example, GRIPS~\citep{prasad2023grips} splits instructions into phrases and applies phrase-level edit operations: delete, swap, paraphrase, and addition, to progressively improve prompt quality. Plum~\citep{pan2024plum} extends GRIPS by incorporating metaheuristic strategies such as simulated annealing, mutation, and crossover. TEMPERA~\citep{zhang2023tempera} further frames the editing process as a reinforcement learning problem, training a policy model to perform different editing techniques to construct query-dependent prompts efficiently. 

\subsubsection{Generative Prompt Optimisation}

In contrast to edit-based methods that apply local modifications to prompts, \textit{generative} approaches leverage LLMs to iteratively generate entirely new prompts, conditioned on a base prompt and various optimisation signals. Compared to local edits, generative methods can explore a broader region of the prompt space and produce more diverse and semantically rich candidates. 

The prompt generation process is typically driven by a variety of optimisation signals that guide the LLM towards producing improved prompts. These signals may include predefined rewriting rules~\citep{xu2022gps, zhou-etal-2024-fairer}, input-output examplars~\citep{zhou2023large,xu2024reprompting}, and dataset or program descriptions~\citep{opsahl2024optimizing}. Additional guidance can come from prior prompts along with their evaluation scores~\citep{yang2024large}, meta-prompts that specify task objectives and constraints~\citep{ye2023prompt,hsieh2024automatic,wang2024promptagent,xiang2025self}, as well as signals that indicate the desired direction of change~\citep{fernando2024promptbreeder,guo2024evoprompt,opsahl2024optimizing}. Moreover, some methods also leverage success and failure examples to highlight effective or problematic prompt patterns~\citep{wu2024strago,yao2024retroformer}. 
For example, ORPO~\citep{yang2024large} generates new instructions by prompting the LLM with previously generated candidates and their evaluation scores. StraGo~\citep{wu2024strago} leverages insights from both successful and failure cases to identify critical factors for obtaining high-quality prompts. The optimisation signals can be further integrated into advanced search strategies, such as Gibbs sampling~\citep{xu2024reprompting}, Monte Carlo tree search (MCTS)~\citep{wang2024promptagent}, Bayesian optimisation~\citep{opsahl2024optimizing,lin2024use,hu2024localized,schneider2025hyperband,wan2025few}, and neural bandit-based methods~\citep{lin2024use,shi2024best,lin2024prompt}. These search strategies enable more efficient and scalable exploration of the prompt space. For instance, PromptAgent~\citep{wang2024promptagent} formulates prompt optimisation as a strategic planning problem and leverages MCTS to efficiently navigate the expert-level prompt space. MIPRO~\citep{opsahl2024optimizing} employs Bayesian optimisation to efficiently search for the optimal combination of instruction candidates and few-shot demonstrations. 

While most generative approaches use a frozen LLM to generate new prompts, recent work has explored the use of reinforcement learning to train a policy model for prompt generation~\citep{deng2022rlprompt,sun2024query,yao2024retroformer,wang2025ragen}. For example, Retroformer~\citep{yao2024retroformer} trains a policy model to iteratively refine prompts by summarising the root cause of previous failed cases.

\subsubsection{Text Gradient-Based Prompt Optimisation} 
In addition to editing and generating prompts directly, a more recent line of work explores the use of \textit{text gradients} to guide prompt optimisation~\citep{pryzant2023automatic,yuksekgonul2024textgrad,wang2024correctly,austin2024grad,mert2025optimizing,tang2025unleashing,zhang2025revolve}. These methods draw inspiration from gradient-based learning in neural networks, but instead of computing numerical gradients over model parameters, they generate natural language feedback, which is referred to as ``text gradient'', that guides how a prompt should be updated to optimise a given objective. Once the text gradient is obtained, the prompt is updated according to the feedback. The key components within such approaches lie in how the text gradients are generated and how they are subsequently used to update the prompt. For example, ProTeGi~\citep{pryzant2023automatic} generates text gradients by criticising the current prompt. Subsequently, it edits the prompt in the opposite semantic direction of the gradient. Such ``gradient descent'' steps are guided by a beam search and bandit selection procedure to find optimal prompts efficiently. Similarly, TextGrad~\citep{yuksekgonul2024textgrad,mert2025optimizing} generalises this idea to a broader framework for compound AI systems. It treats textual feedback as a form of “automatic differentiation” and uses LLM-generated suggestions to iteratively improve components such as prompts, code, or other symbolic variables. Another work~\citep{zhou2024symbolic} proposes agent symbolic learning, a data-centric framework that models language agents as symbolic networks and enables them to autonomously optimise their prompts, tools, and workflows via symbolic analogues of back-propagation and gradient descent. Recent work~\citep{wu2025optimas} also explores the prompt optimisation in compound AI systems, where its goal is to automatically optimise the configuration across a heterogeneous set of components and parameters, e.g., model parameters, prompts, model selection choice, and hyperparameters.

\subsubsection{Evolutionary Prompt Optimisation}
In addition to the above optimisation techniques, \textit{evolutionary algorithms} have also been explored as a flexible and effective approach for prompt optimisation~\citep{guo2024evoprompt,fernando2024promptbreeder}. These approaches treat prompt optimisation as an evolutionary process, maintaining a population of candidate prompts that are iteratively refined through evolutionary operators such as mutation, crossover, and selection. For example, EvoPrompt~\citep{guo2024evoprompt} leverages two widely used evolutionary algorithms: Genetic Algorithm (GA) and Differential Evolution (DE), to guide the optimisation process to find the high-performing prompts. It adapts the core evolutionary operations, namely mutation and crossover, to the prompt optimisation setting, where new candidate prompts are generated by combining segments from two parent prompts and introducing random alternation to specific elements. Similarly, Promptbreeder~\citep{fernando2024promptbreeder} also iteratively mutates a population of task-prompts to evolve these prompts. A key feature is its use of \textit{mutation prompts}, which are instructions that specify how task-prompts should be modified during the mutation process. These mutation prompts can be either predefined or generated dynamically by the LLM itself, enabling a flexible and adaptive mechanism for guiding prompt evolution.

\subsection{Memory Optimisation}
Memory is essential for enabling agents to reason, adapt, and operate effectively over extended tasks. However, AI agents frequently face limitations arising from constrained context windows and forgetting, which can result in phenomena such as context drift and hallucination~\citep{liu2023lostmiddlelanguagemodels, zhang2024chain, zhang2024survey}. These limitations have driven increasing interest in memory optimisation to enable generalisable and consistent behaviours in dynamic environments. In this survey, we focus on inference-time memory strategies that enhance memory utilisation without modifying model parameters. In contrast to training-time techniques such as fine-tuning or knowledge editing~\citep{decao2021editingfactualknowledgelanguage,mitchell2022fastmodeleditingscale}, inference-time approaches dynamically decide what to retain, retrieve, and discard during the reasoning process.

We categorise existing methods into two optimisation objectives: \textit{short-term memory}, which focuses on maintaining coherence within the active context, and \textit{long-term memory}, which supports persistent retrieval across sessions. This optimisation-oriented perspective shifts the focus from static memory formats (e.g., internal vs. external) to dynamic memory control, with an emphasis on how memory is scheduled, updated, and reused to support decision-making. In the following subsections, we present representative methods within each category, emphasising their impact on reasoning fidelity and effectiveness in long-horizon settings.

\subsubsection{Short-term Memory Optimisation}
Short-term memory optimisation focuses on managing limited contextual information within the LLM’s working memory~\citep{liu2023lostmiddlelanguagemodels}. This typically includes recent dialogue turns, intermediate reasoning traces, and task-relevant content from the immediate context. As the context expands, memory demands increase significantly, making it impractical to retain all information within a fixed context window. To address this, various techniques have been proposed to compress, summarise, or selectively retain key information~\citep{zhang2024survey, wang2025recursivelysummarizingenableslongterm}. Common strategies encompass summarisation, selective retention, sparse attention, and dynamic context filtering. For example, \cite{wang2025recursivelysummarizingenableslongterm} proposes recursive summarisation to incrementally construct compact and comprehensive memory representations, enabling consistent responses throughout extended interactions. MemoChat~\citep{lu2023memochattuningllmsuse} maintains dialogue-level memory derived from conversation history to support coherent and personalised interaction. COMEDY~\citep{chen2024compressimpressunleashingpotential} and ReadAgent~\citep{lee2024humaninspiredreadingagentgist} further incorporate extracted or compressed memory traces into the generation process, allowing agents to maintain context over long conversations or documents. In addition to summarisation, other methods dynamically adjust the context or retrieve intermediate state traces to facilitate multi-hop reasoning. For example, MoT~\citep{li2023motmemoryofthoughtenableschatgpt} and StructRAG~\citep{li2024structragboostingknowledgeintensive} retrieve self-generated or structured memory to guide intermediate steps. MemoryBank~\citep{zhong2023memorybankenhancinglargelanguage}, inspired by the Ebbinghaus forgetting curve~\citep{murre2015replication}, hierarchically summarises events and updates memory based on recency and relevance. Reflexion~\citep{shinn2023reflexionlanguageagentsverbal} enables agents to reflect on task feedback and store episodic insights, promoting self-improvement over time. 

These methods significantly improve local coherence and context efficiency. However, short-term memory alone is insufficient for retaining knowledge across sessions or enabling generalisation over long horizons, highlighting the need for complementary long-term memory mechanisms.

\subsubsection{Long-term Memory Optimisation}
Long-term memory optimisation mitigates the limitations of short context windows by providing persistent and scalable storage that extends beyond the immediate input scope of the language model. It enables agents to retain and retrieve factual knowledge, task histories, user preferences, and interaction trajectories across sessions~\citep{du2025rethinkingmemoryaitaxonomy}, thereby supporting coherent reasoning and decision-making over time. A key objective in this area is to manage increasingly complex and expanding memory spaces while preserving a clear separation between memory storage and the reasoning process~\citep{zhang2024survey}. External memory can be either unstructured or organised into structured formats such as tuples, databases, or knowledge graphs~\citep{zeng2024structuralmemoryllmagents}, and may span a wide range of sources and modalities.

A critical paradigm of long-term memory optimisation is Retrieval-Augmented Generation (RAG), which incorporates relevant external memory into the reasoning process via retrieval~\citep{wang2023voyageropenendedembodiedagent,efeoglu2024retrieval,gao2025synergizingragreasoningsystematic}. For instance, EWE~\citep{chen2025improvingfactualityexplicitworking} augments a language model with an explicit working memory that dynamically holds latent representations of retrieved passages, focusing on combining static memory entries at each decoding step. In contrast, A-MEM~\citep{xu2025amemagenticmemoryllm} constructs interconnected knowledge networks through dynamic indexing and linking, enabling agents to form evolving memory. Another prominent direction involves agentic retrieval, where agents autonomously determine when and what to retrieve, alongside trajectory-level memory, which utilises past interactions to inform future behaviour. Supporting techniques such as efficient indexing, memory pruning, and compression further enhance scalability~\citep{zheng2024synapsetrajectoryasexemplarpromptingmemory,alizadeh2024llmflashefficientlarge}. For example, \cite{wang2024machineunlearningmeetsretrievalaugmented} propose a lightweight unlearning framework based on the RAG paradigm. By altering the external knowledge base used for retrieval, the system can simulate forgetting effects without modifying the underlying LLM. Similarly, \cite{xu2025amemagenticmemoryllm} introduce a self-evolving memory system that maintains long-term memory without relying on predefined operations. In addition to retrieval policies and memory control mechanisms, the structure and encoding of memory itself significantly affect system performance. Vector-based memory systems encode memory in dense latent spaces and support fast, dynamic access. For instance, MemGPT~\citep{packer2024memgptllmsoperatingsystems}, NeuroCache~\citep{safaya-yuret-2024-neurocache}, G-Memory~\citep{zhang2025g}, and AWESOME~\citep{cao2023awesomegpumemoryconstrainedlong} enable consolidation and reuse across tasks. Mem0~\citep{chhikara2025mem0buildingproductionreadyai} further introduces a production-ready memory-centric architecture for continuous extraction and retrieval. Other approaches draw inspiration from biological or symbolic systems to improve interpretability. HippoRAG~\citep{gutiérrez2025hipporagneurobiologicallyinspiredlongterm} implements hippocampus-inspired indexing via lightweight knowledge graphs. GraphReader~\citep{li2024graphreaderbuildinggraphbasedagent} and Mem0\textsuperscript{g}~\citep{chhikara2025mem0buildingproductionreadyai} use graph-based structures to capture conversational dependencies and guide retrieval. In the symbolic domain, systems like ChatDB~\citep{hu2023chatdbaugmentingllmsdatabases} issue SQL queries over structured databases, while \cite{wang2024symbolicworkingmemoryenhances} introduces a neurosymbolic framework that stores facts and rules in both natural and symbolic form, supporting precise reasoning and memory tracking.

Recent work has also emphasised the importance of memory control mechanisms during inference~\citep{10756271, chen2025improvingfactualityexplicitworking}, which determine what, when, and how to store, update, or discard memory~\citep{jin2025disentanglingmemoryreasoningability}. For instance, MATTER~\citep{lee-etal-2024-matter} dynamically selects relevant segments from multiple heterogeneous memory sources to support question answering, and AWM~\citep{wang2024agentworkflowmemory} enables continuous memory updates in both online and offline settings. MyAgent~\citep{Hou_2024} endows agents with memory-aware recall mechanisms for generation, addressing the temporal cognition limitations of LLMs. MemoryBank~\citep{zhong2023memorybankenhancinglargelanguage} proposes a cognitively inspired update strategy, where periodic revisiting of past knowledge mitigates forgetting and enhances long-term retention. Reinforcement learning and prioritisation policies have also been employed to guide memory dynamics~\citep{zhou2025agentfly,yan2025memory,long2025seeing}. For example, MEM1~\citep{zhou2025mem1learningsynergizememory} leverages reinforcement learning to maintain an evolving internal memory state, selectively consolidating new information while discarding irrelevant content. A-MEM~\citep{xu2025amemagenticmemoryllm} presents an agentic memory architecture that autonomously organises, updates, and prunes memory based on usage. MrSteve~\citep{park2025mrsteveinstructionfollowingagentsminecraft} incorporates episodic ``what-where-when'' memory to hierarchically structure long-term knowledge, enabling goal-directed planning and task execution. These approaches allow agents to proactively manage memory and complement short-term mechanisms. Meanwhile, MIRIX~\citep{wang2025mirixmultiagentmemoryllmbased} introduces an agent memory system with six specialised memory types in collaborative settings, enabling coordinated retrieval and achieving state-of-the-art performance in long-horizon tasks, while Agent KB~\citep{tang2025agent} leverages a shared knowledge base with a teacher-student dual-phase retrieval mechanism to transfer cross-domain problem-solving strategies and execution lessons across agents, significantly enhancing performance through hierarchical strategic guidance and refinement.

\subsection{Tool Optimisation}
Tools are critical components within agent systems, serving as interfaces that allow agents to perceive and interact with the real world. They enable access to external information sources, structured databases, computational resources, and APIs, thereby enhancing the agent's ability to solve complex, real-world problems~\citep{patil2024gorilla, yang2023gpt4tools, guo2024stabletoolbench}. As a result, tool use has become a core competence of AI agents, especially for tasks that require external knowledge and multi-step reasoning. However, simply exposing agents to tools is not sufficient. Effective tool use requires the agent to recognise when and how to invoke the right tools, interpret tool outputs, and integrate them into multi-step reasoning. Consequently, recent research has focused on tool optimisation, which aims to enhance the agent's ability to use tools intelligently and efficiently. 

Existing research on tool optimisation largely falls into two complementary directions. The first, which has been more extensively explored, focuses on enhancing the agent's ability to interact with tools. This is achieved through different approaches, including {training strategies, prompting techniques, and reasoning algorithms}, that aim to improve the agent's ability to understand, select, and execute tools effectively. 
The second, which is more recent and still emerging, focuses on optimising the tools themselves by modifying existing tools or creating new ones that better align with the functional requirements of the target tasks. 

\subsubsection{Training-Based Tool Optimisation} 
Training-based tool optimisation aims to enhance an agent's ability to use tools by updating the underlying LLM's parameters through learning. The motivation behind this approach stems from the fact that LLMs are pretrained purely on text generation tasks, without any exposure to tool usage or interactive execution. Therefore, they lack an inherent understanding of how to invoke external tools and interpret tool outputs. Training-based methods aim to address this limitation by explicitly teaching the LLMs how to interact with tools, thereby embedding tool-use capabilities directly into the agent’s internal policy. 

\paragraph{{Supervised Fine-Tuning for Tool Optimisation.}}
Earlier efforts in this line of work rely on supervised fine-tuning (SFT), which trains the LLM on high-quality tool-use trajectories to explicitly demonstrate how tools should be invoked and integrated into task execution~\citep{schick2023toolformer,du2024anytool,liu2025toolace,wang2025toolgen}. A central focus of these methods lies in the collection of high-quality tool-use trajectories, which typically consist of input queries, intermediate reasoning steps, tool invocations and final answers. These trajectories serve as explicit supervision signals for the agent, teaching it how to plan tool usage, execute calls, and incorporate results into its reasoning process. For example, approaches such as ToolLLM~\citep{qin2024toolllm} and GPT4Tools~\citep{yang2023gpt4tools} leverage more powerful LLMs to generate both instructions and corresponding tool-use trajectories. Inspired by the human learning process, STE~\citep{wang2024llms} introduces simulated trial-and-error interactions to collect tool-use examples, while TOOLEVO~\citep{chenlearning} employs MCTS to enable more active exploration and collect higher-quality trajectories. T3-Agent~\citep{gao2025multi} further extends this paradigm to the multimodal setting by introducing a data synthesis pipeline that generates and verifies high-quality multimodal tool-use trajectories for tuning vision–language models. 

Moreover, recent work~\citep{yao2025taubench} indicates that even advanced LLMs face challenges with tool use in multi-turn interactions, especially when these interactions involve complex function calls, long-term dependencies, or requesting missing information. To generate high-quality training trajectories on multi-turn tool calling, Magnet~\citep{yin2025magnet} proposes to synthesise a sequence of queries and executable function calls from tools, and employs a graph to build a reliable multi-turn query. BUTTON~\citep{chen2025facilitating} generates synthetic compositional instruction tuning data via a two-stage process, where a bottom-up stage composes atomic tasks to construct the instructions and a top-down stage employs a multi-agent system to simulate the user, assistant, and tool to generate the trajectory data. 
To enable more realistic data generation, APIGen-MT~\citep{prabhakar2025apigen} proposes a two-phase framework that first generates tool call sequences and then transforms them into complete multi-turn interaction trajectories through simulated human-agent interplay. 

Once the tool-use trajectories are collected, they are used to fine-tune the LLM through standard language modelling objectives, enabling the model to learn successful patterns of tool invocation and integration. In addition to this common paradigm, some studies have also explored more advanced training strategies to further enhance tool-use capabilities. For example, Confucius~\citep{gao2024confucius} introduces an easy-to-difficult curriculum learning paradigm that gradually exposes the model to increasingly complex tool-use scenarios. Gorilla~\citep{patil2024gorilla} proposes integrating a document retriever into the training pipeline, allowing the agent to dynamically adapt to evolving toolsets by grounding tool usage in retrieved documentation. 

\paragraph{Reinforcement Learning for Tool Optimisation.} 
While supervised fine-tuning has proven effective for teaching agents to use tools, its performance is often constrained by the quality and coverage of the training data. Low-quality trajectories can lead to diminished performance gains. Moreover, fine-tuning on limited datasets may hinder generalisation, particularly when agents encounter unseen tools or task configurations at inference time. To address these limitations, recent research has turned to reinforcement learning (RL) as an alternative optimisation paradigm for tool use. By enabling agents to learn through interaction and feedback, RL facilitates the development of more adaptive and robust tool-use strategies. This approach has shown promising results in recent work such as ReTool~\citep{feng2025retool} and Nemotron-Research-Tool-N1 (Tool-N1)~\citep{zhang2025nemotron}, both of which demonstrate how lightweight supervision in an interactive environment can lead to more generalisable tool-use capabilities. Tool-Star~\citep{dong2025tool} enhances RL-based tool use by combining scalable tool-integrated data synthesis with a two-stage training framework to improve autonomous multi-tool collaborative reasoning. SPORT~\citep{li2025iterative} extends RL-based tool optimisation to the multimodal setting through step-wise preference optimisation, enabling agents to self-synthesise tasks, explore and verify tool usage without human annotations. 
Building on these foundations, further studies have focused on improving RL algorithms for tool use, including ARPO~\citep{dong2025agentic}, which balances long-horizon reasoning and multi-turn tool interactions via an entropy-based adaptive rollout mechanism and stepwise advantage attribution, as well as methods that design more effective reward functions~\citep{qian2025toolrl} and leverage synthetic data generation and filtering to enhance training stability and efficiency~\citep{goldie2025synthetic}. 

\subsubsection{Inference-Time Tool Optimisation}
In addition to training-based approaches, another line of work focuses on enhancing tool-use capabilities during inference, without modifying LLM parameters. These methods typically operate by optimising tool-related contextual information within prompts or guiding the agent's decision-making process through structured reasoning at test time. There are two major directions within this paradigm: (1) \textit{prompt-based methods}, which refine the representation of tool documentation or instructions to facilitate better understanding and utilisation of tools; (2) \textit{reasoning-based methods}, which leverage test-time reasoning strategies, such as MCTS and other tree-based algorithms to enable more effective exploration and selection of tools during inference. 

\paragraph{Prompt-Based Tool Optimisation.}
Tool-related information is typically provided to agents through tool documentation within prompts. These documents describe tool functionalities, potential usage, and invocation formats, helping the agent understand how to interact with external tools to solve complex tasks.  Therefore, tool documentation within prompts serves as a crucial bridge between the agent and its available tools, directly influencing the quality of tool-use decisions. Recent efforts have focused on optimising how this documentation is presented, either by restructuring source documents or refining them through interactive feedback~\citep{qu2025from}. For instance, EASYTOOL~\citep{yuan-etal-2025-easytool} transforms different tool documentation into unified, concise instructions, making them easier for LLMs to use. In contrast, approaches such as DRAFT~\citep{qu2025from} and PLAY2PROMPT~\citep{fang2025play2prompt} draw inspiration from human trial-and-error processes, introducing interactive frameworks that iteratively refine tool documentation based on feedback. 

Beyond these methods, a more recent direction explores the joint optimisation of both tool documentation and the instructions provided to the LLM agent. For example, \cite{wu2025joint} propose an optimisation framework that simultaneously refines the agent's prompt instructions and the tool descriptions, collectively referred to as the \textit{context}, to enhance their interaction. The optimised context has been shown to reduce computational overhead and improve tool-use efficiency, highlighting the importance of context design in effective inference-time tool optimisation. 

\paragraph{Reasoning-Based Tool Optimisation.} 
Test-time reasoning and planning techniques have demonstrated strong potential for improving tool-use capabilities in AI agents. Early work such as ToolLLM~\citep{qin2024toolllm} has validated the effectiveness of the ReAct \citep{yao2023react} framework in tool-use scenarios, and further proposed a depth-first tree search algorithm that enables agents to quickly backtrack to the last successful state rather than restarting from scratch, which significantly improves efficiency. 
ToolChain~\citep{zhuang2024toolchain} introduces a more efficient tree-based search algorithm by employing a cost function to estimate the future cost of a given branch. This allows agents to prune low-value paths early and avoid the inefficient rollouts commonly associated with traditional MCTS. Similarly, Tool-Planner~\citep{liu2025toolplanner} clusters tools with similar functionalities into \textit{toolkits} and leverages a tree-based planning method to quickly reselect and adjust tools from these toolkits. MCP-Zero~\citep{fei2025mcp} introduces an active agent framework that empowers LLMs to autonomously identify capability gaps and request tools on demand.  

\subsubsection{Tool Functionality Optimisation}
Beyond optimising the agent's behaviour, a complementary line of work focuses on modifying or generating tools themselves to better support task-specific reasoning and execution. Inspired by the human practice of continuously developing tools to meet task requirements, these approaches aim to extend the agent's action space by adapting the toolset to the task, rather than adapting the task to a fixed toolset~\citep{wang2024trove}. For instance, CREATOR~\citep{qian2023creator} and LATM~\citep{cai2024large} introduce frameworks that generate tool documentation and executable code for novel tasks. CRAFT~\citep{yuan2024craft} leverages reusable code snippets from prior tasks to create new tools for unseen scenarios. AgentOptimiser~\citep{zhang2024offline} treats tools and functions as learnable weights, allowing the agent to iteratively refine them using LLM-based updates. A more recent work, Alita~\citep{qiu2025alita}, extends tool creation into a Multi-Component Program (MCP) format, which enhances reusability and environment management. Moreover, CLOVA~\citep{gao2024clova} introduces a closed-loop visual assistant framework with inference, reflection, and learning phases, enabling continual adaptation of visual tools based on human feedback. 

\section{Multi-Agent Optimisation} 
\label{sec:multi_agent_optimisation}


The multi-agent workflow defines how multiple agents collaborate to solve complex tasks through structured topologies and interaction patterns. The field has witnessed a fundamental shift: from manually designed agent architectures, where researchers explicitly specify collaboration patterns and communication protocols, to self-evolving systems that automatically discover effective collaboration strategies. This evolution reframes workflow design as a search problem over three interconnected spaces: the structural space of possible agent topologies, the semantic space of agent roles and instructions, and the capability space of LLM backbones. Recent approaches explore these spaces using a range of optimisation techniques, from evolutionary algorithms to reinforcement learning, each offering different trade-offs in balancing multiple optimisation targets (e.g., accuracy, efficiency, and safety).

\begin{figure}[H]
  \centering
  \includegraphics[width=0.7\linewidth]{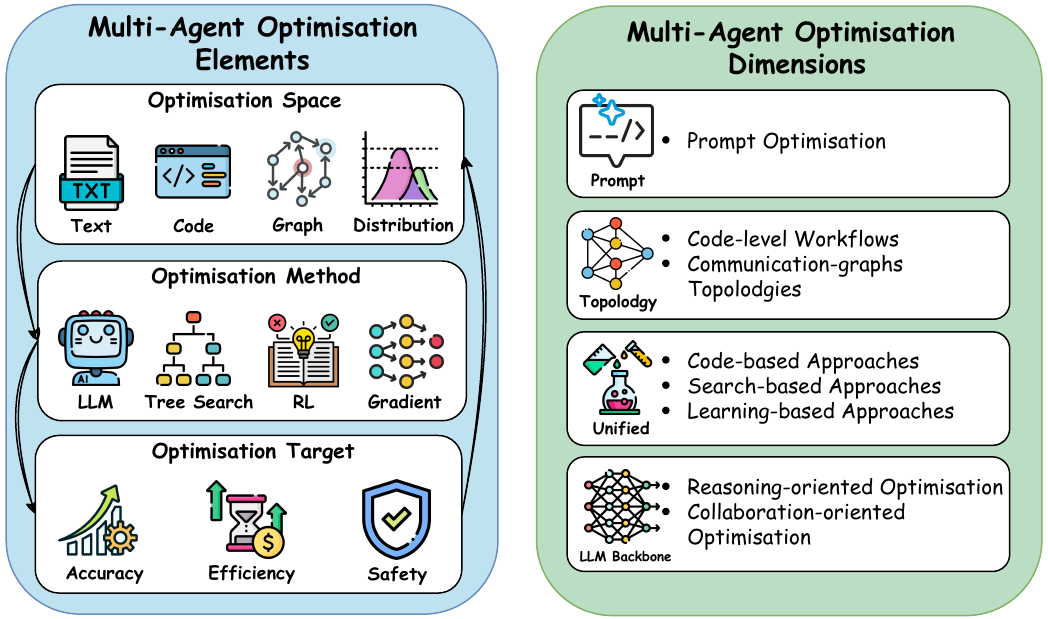}
      \caption{An overview of multi-agent systems optimisation approaches, with core optimisation elements (space, methods, and targets) on the left and optimisation dimensions (prompt, topology, unified, and LLM backbone) on the right.}
  \label{fig:workflow}
\end{figure}

This section traces the progression of multi-agent workflow optimisation across four key dimensions. Our starting point examines manually designed paradigms that establish foundational principles. From there, we consider prompt-level optimisation, which refines agent behaviours within fixed topologies. We subsequently address topology optimisation, which focuses on discovering the most effective architectures for multiple agents to accomplish a given task. We also discuss comprehensive approaches that simultaneously consider multiple optimisation spaces, jointly optimising prompts, topologies, and other system parameters in an integrated manner. Additionally, we investigate LLM-backbone optimisation, which enhances the fundamental reasoning and collaborative capabilities of the agents themselves through targeted training. Through this lens, we show how the field progressively expands its conception of what constitutes a searchable and optimisable parameter in multi-agent systems, from agent instructions and communication structures to the core competencies of the underlying models. Figure~\ref{fig:workflow} provides an overview of multi-agent workflow optimisation across its core elements and key dimensions.

\subsection{Manually Designed Multi-Agent Systems}Manually designed workflows form the foundation of multi-agent collaboration research. These architectures encode researchers' insights about task decomposition, agent capabilities, and coordination mechanisms into explicit interaction patterns. By examining these handcrafted paradigms, we can understand the design principles that guide agent collaboration and the engineering considerations that shape system architecture. 

\paragraph{Parallel Workflows.} Parallel workflows employ concurrent execution followed by collective decision-making. The simplest form involves multiple independent agents generating solutions in parallel, followed by majority voting to select the final output. Empirical evidence shows that parallel generation with small LLMs can match or even outperform single large LLMs~\citep{anonymous2024replacing, wang2025mixtureofagents}. Multi-layer aggregation further reduces error bounds and improves robustness~\citep{zhang2025optimizingsequentialmultisteptasks}. Recent extensions incorporate dynamic task graphs and asynchronous threads to enable near-linear scaling and lower decision latency~\citep{yu2025dyntaskmasdynamictaskgraphdriven,gu2025agentgroupchatv2divideandconquerllmbasedmultiagent,wang2025all}. However, while computational throughput scales horizontally, the engineering costs of managing coordination and consistency grow exponentially. 

\paragraph{Hierarchical Workflows.}
When subtasks exhibit strict contextual dependencies, hierarchical~\citep{zhang2024chain,qian2024chatdevcommunicativeagentssoftware} workflows offer a structured alternative. These frameworks organise agents into multi-level top-down structures or sequential pipelines. The system decomposes tasks across layers, with each layer responsible for different subtasks. This design excels in complex goal-driven tasks such as deep research and code generation~\citep{hong2023metagpt,zhang2025agentorchestrahierarchicalmultiagentframework}. However, its fixed topology limits adaptability, especially when facing dynamic goals or resource constraints.

\paragraph{Multi-Agent Debate.} 
To balance accuracy with interpretability, researchers have developed the debate paradigm, where agents engage in adversarial-negotiation-arbitration cycles to discuss and correct reasoning errors. Early work explored symmetric debater mechanisms~\citep{li2024improvingmultiagentdebatesparse}. 
More recent studies extend this framework by introducing role asymmetry, adjustable debate intensity, and persuasiveness-oriented strategies~\citep{yin-etal-2023-exchange,liang-etal-2024-encouraging,khan2024debating,chang2024socrasynthmultillmreasoningconditional}. In addition, confidence-gated debate strategies demonstrate that triggering multi-agent debates only when a single model exhibits low confidence can sharply reduce inference costs without hindering performance~\citep{eo2025debatenecessaryadaptivemultiagent}.

Despite the success of manually designed workflows and structured multi-agent paradigms, recent empirical studies reveal that single large LLMs with well-crafted prompts can match the performance of complex multi-agent discussion frameworks on multiple reasoning benchmarks~\citep{pan2025why}. This finding, coupled with the high implementation and maintenance costs of handcrafted multi-agent workflows~\citep{li2024autoflowautomatedworkflowgeneration,zhang2025aflow}, has driven the development of \textit{self-evolving multi-agent systems} that can automatically learn, adapt, and restructure their workflows over time, rather than relying on fixed architectures and static coordination protocols.

\subsection{Self-Evolving Multi-Agent System}
The high engineering costs and limited adaptability of manually designed multi-agent workflows have motivated a shift towards automated, self-evolving systems. These systems can automatically design, evaluate, and refine agent workflows by adapting their prompts, topologies, and collaborative strategies based on performance feedback. Instead of relying on hard-coded configurations, they treat workflow optimisation as a search problem, where the system explores and optimises over a space of possible configurations. The search space spans multiple levels, from local prompts to global topology structures. 

To effectively navigate the search space, various search algorithms have been introduced. 
These methods range from reinforcement learning, Monte Carlo Tree Search, and generative models that enable efficient exploration, to evolutionary operators that provide robust search capabilities.
Moreover, the optimisation objectives have expanded from improving performance to balancing multi-dimensional goals, including task accuracy, computational efficiency, and safety. This evolution reveals that as search capabilities advance, the core challenge shifts from finding optimal solutions to defining what optimality means in dynamic multi-agent contexts.

\subsubsection{Multi-Agent Prompt Optimisation}
One promising direction for achieving such self-evolution is through prompt optimisation, where prompts define both agent roles and their corresponding task instructions. Recent approaches treat these prompt-encoded configurations as a formal search space for systematic refinement. In fact, prompt optimisation in multi-agent workflows often builds upon the single-agent techniques discussed in Section~\ref{subsec:prompt_optimisation}, but extends them to coordinate multiple agents and task dependencies. 
For example, 
DSPy~\citep{DBLP:journals/corr/abs-2312-13382} Assertions introduces runtime self-evolution, where the search space encompasses possible intermediate outputs from pipeline modules, using assertion-driven backtracking with explicit feedback to guide LLMs in self-correcting outputs that violate programmatic constraints.
AutoAgents~\citep{chen2024autoagentsframeworkautomaticagent} extends prompt optimisation from single-agent settings to entire multi-agent team configurations, optimising specialised agent roles and execution plans through structured dialogue between dedicated meta-agents.

\subsubsection{Topology Optimisation}

Topology optimisation represents a paradigm shift in multi-agent system design: rather than treating communication structure as a fixed constraint, it recognises topology itself as a powerful optimisation target. This insight emerged from a fundamental observation—even the best prompts cannot compensate for poor architectural choices. Viewed through a representation-centred lens, existing work clusters into two complementary families: program/code-level workflow topologies and communication-graph topologies; this classification foregrounds \emph{what} is being optimised—the chosen representation of topology. This marks not just technical progress but a conceptual shift—the medium (topology) matters as much as the message (prompts).

\paragraph{{Code-level workflows.}}
Representing workflows as executable programs or typed code graphs makes agent coordination explicit and verifiable, enabling compositional reuse and automated checking. AutoFlow~\citep{li2024autoflowautomatedworkflowgeneration} sets the search space to natural-language programs (CoRE) and trains a generator LLM with reinforcement learning, supporting both fine-tuning and in-context use. 
Compared with AutoFlow, AFlow~\citep{zhang2025aflow}  replaces the NL program space with typed, reusable operators to form code graphs; Monte Carlo Tree Search with LLM\mbox{-}guided expansion and soft probabilistic selection provides a more structured, sample\mbox{-}efficient exploration of the vast design space than RL over CoRE.
Pushing beyond these discrete search schemes, ScoreFlow~\citep{wang2025scoreflow} lifts code representations into a continuous space and applies gradient\mbox{-}based optimisation with Score\mbox{-}DPO (a direct preference optimisation variant incorporating quantitative feedback) to improve the workflow generator. This addresses the exploration inefficiency inherent to RL/MCTS and enables task\mbox{-}level adaptive workflow generation.
Orthogonal to search\mbox{-}based optimisation, MAS\mbox{-}GPT~\citep{ye2025masgpttrainingllmsbuild} uses supervised fine\mbox{-}tuning on a consistency\mbox{-}oriented corpus (inter\mbox{-} and intra\mbox{-}consistency) so that a single inference aims to produce a complete, executable MAS codebase, trading broad search coverage for one\mbox{-}shot efficiency and stronger dependence on data quality.

\paragraph{{Communication-graph topologies.}}
Unlike code\mbox{-}level programs, this line treats the workflow as a multi-agent communication graph whose connections are the optimisation target~\citep{liu2025graphaugmentedlargelanguagemodel}. 
GPTSwarm~\citep{zhuge2024gptswarm} defines its search space as connections within a computational graph of agents. It relaxes this discrete space into continuous edge probabilities, also employing RL to learn optimal connection schemes.
Building on GPTSwarm, DynaSwarm~\citep{leong2025dynaswarmdynamicallygraphstructure} extends the search space from a single optimised graph to a portfolio of graph structures with Actor–Critic (A2C) optimisation and a lightweight graph selector for per-instance topology selection, addressing the key observation that different queries require different graph structures for optimal performance.
Rather than masking edges in a fixed space, G\mbox{-}Designer~\citep{zhang2025gdesignerarchitectingmultiagentcommunication} employs a variational graph autoencoder to directly generate task-adaptive communication graphs, modulating structural complexity to balance quality and token cost. 
MermaidFlow~\citep{zheng2025mermaidflowredefiningagenticworkflow} represents topology as a typed, declarative graph with static verification and explores only semantically valid regions via safety-constrained evolutionary operators.
 
Beyond static graph synthesis, some approaches dynamically modulate the communication graph during execution.
DyLAN~\citep{liu2023dynamic} treats the search space as active agents across layers with an early-stopping time axis; it prunes low-value agents via an LLM ranker and performs automated team optimisation with an Agent Importance Score using propagation–aggregation–selection.
Captain Agent~\citep{song2025adaptiveinconversationteambuilding} defines the search space as subtask-specific sets of agents and tools (retrieved, filtered, and, when needed, generated); nested group conversations and reflection iteratively refine team composition in situ rather than synthesising a fixed graph from scratch.
Flow~\citep{niu2025flow} contrasts with DyLAN's pruning and Captain Agent's team recomposition by dynamically adjusting the AOV graph structure: it selects an initial graph via parallelism/dependency metrics and then refines it online through workflow refinement and subtask reassignment, achieving modular concurrency with minimal coordination cost.

Orthogonal to graph synthesis, pruning methods optimise by removing redundant or risky communications while preserving essential collaboration.
AgentPrune~\citep{zhang2025cut} treats the search space as a spatial-temporal communication graph where both intra-dialogue (spatial) and inter-dialogue (temporal) edges are pruning targets; it employs a trainable low-rank-guided graph mask to identify and eliminate redundant communications via one-shot pruning, optimizing for token economy.
Building on this pruning paradigm, AGP (Adaptive Graph Pruning)~\citep{li2025adaptivegraphpruningmultiagent} extends the search space to include both agent quantity (hard pruning) and communication edges (soft pruning). It employs a two-stage training strategy that jointly optimises these dimensions on a per-task basis, dynamically determining the optimal number of agents and their connections for task-specific topology generation.
While the above methods prune for efficiency and adaptability, G-Safeguard~\citep{wang2025gsafeguardtopologyguidedsecuritylens} applies pruning for security—it operates on communication edges as the search space, using a GNN to flag risky nodes and deterministic rules to cut outward edges under a model-driven threshold for defence against adversarial attacks. Relatedly, NetSafe~\citep{yu2024netsafeexploringtopologicalsafety} summarises topological safety risks and proposes graph-based detection and intervention principles as a complementary safety lens.

\subsubsection{Unified Optimisation}
Unified optimisation emerges from a key insight: prompts and topology are not independent design choices but deeply interconnected aspects of agent systems~\citep{zhou2025multi}. A well-crafted prompt cannot function effectively in a poor communication structure, while an elegant topology yields little benefit with poorly instructed agents. This interdependence has driven the field along three distinct technical paths: code-based unification, structured optimisation methods, and learning-driven architectures. Each approach tackles the joint optimisation challenge from a unique angle, revealing different trade-offs between efficiency and performance.

\paragraph{{Code-based Approaches.}}
The most direct approach to unified optimisation treats code as a universal representation for both prompts and topology. ADAS~\citep{hu2025automated} pioneered this approach through its Meta Agent Search framework, representing prompts, workflows, and tool use as Python code to enable iterative agent generation and evaluation. This code-centric view allows natural co-evolution, modifying agent logic inherently affects both instructional and structural aspects. FlowReasoner~\citep{zhang2025evoflowevolvingdiverseagentic} advanced the code-based paradigm by focusing on query-level adaptation, generating one MAS per query rather than per task. After distilling reasoning abilities from DeepSeek-R1, it employs GRPO with external execution feedback to enhance its meta-agent, optimising for performance and efficiency. Together, these methods demonstrate that code provides a flexible substrate for joint optimisation, though at different granularities of adaptation.

\paragraph{{Search-based Approaches.}}
Rather than relying on implicit co-evolution through code, another line of work develops explicit mechanisms for coordinating prompt and topology design. EvoAgent~\citep{yuan-etal-2025-evoagent} defined search spaces as textual agent settings (roles, skills, prompts) and employed evolutionary algorithms with mutation, crossover, and selection operators to generate diverse agent populations. Compared with implicit code\mbox{-}based co\mbox{-}evolution, EvoAgent explicitly evolves configuration\mbox{-}level characteristics rather than synthesising programs.
Relative to EvoAgent’s text-centric configuration search, EvoFlow~\citep{zhang2025evoflowevolvingdiverseagentic} likewise adopts evolutionary search but over operator\mbox{-}node workflow graphs. It introduces predefined composite operators (e.g. CoT, debate) and uses an operator library with tag selection to constrain mutation/crossover and narrow the search space. EvoFlow further treats LLM selection as a decision variable to balance performance and cost; diversity\mbox{-}aware selection preserves population variety, and a multi\mbox{-}objective fitness drives cost–performance Pareto optimisation.

Complementary to evolutionary searches, 
MASS~\citep{zhou2025multi} proposes a three-stage, conditionally coupled optimisation framework: it first locally tunes each agent’s prompts, then searches the workflow topology in a pruned space, and finally performs global prompt optimisation on the selected topology; the procedure alternates rather than fully decoupling, serving as a practical approximation to joint optimisation. 
Most recently, DebFlow~\citep{su2025debflowautomatingagentcreation} represents search spaces as workflow graphs of operator nodes and employs multi-agent debate for optimisation. Guided by reflexion on execution failures, it avoids exhaustive search while pioneering debate mechanisms in automated agent design. These structured approaches trade some flexibility for more targeted optimisation strategies. 
Building on the operator node representation, 
MAS-ZERO~\citep{ke2025maszerodesigningmultiagentsystems} casts unified optimisation as a purely inference-time search, iteratively restructuring agent teams and task decompositions through solvability-guided refinement without any gradient updates or offline training.

\paragraph{{Learning-based Approaches.}}
The latest wave of research applies sophisticated learning paradigms to jointly optimise prompts and topology. MaAS~\citep{zhang2025multiagentarchitecturesearchagentic} shifts from optimising single architectures to learning agentic supernets—probabilistic distributions over multi-agent systems. Its controller network samples query-specific architectures with Monte Carlo and textual gradient optimisation, achieving superior performance with dramatically reduced inference costs. ANN~\citep{ma2025agenticneuralnetworksselfevolving} conceptualises multi-agent collaboration as layered neural networks, where each layer forms specialised agent teams. It employs a two-phase optimisation process: forward task decomposition and backward textual gradient refinement. This approach jointly evolves agent roles, prompts, and inter-layer topologies, enabling post-training adaptation to novel tasks.

\subsubsection{LLM Backbone Optimisation} The evolution of the LLM backbone behind agents is a critical aspect of multi-agent evolution, particularly how agents improve their \textit{cooperative} or \textit{reasoning} abilities through interaction.

\paragraph{{Reasoning-oriented Optimisation.}} A prominent line of work focuses on enhancing the backbone LLM's reasoning capacity via multi-agent collaboration. For instance, \emph{multi-agent finetuning}~\citep{subramaniam2025multiagent-ft} leverages high-quality cooperative trajectories sampled from multi-agent debates for supervised fine-tuning, enabling (1) role-specific specialisation of agents and (2) improved reasoning capabilities of the underlying backbone model. Similarly, Sirius~\citep{zhao2025sirius} and MALT~\citep{motwani2024malt} employ self-play to collect high-quality cooperative trajectories and train agents within their respective multi-agent collaboration frameworks. While both approaches leverage failed trajectories to some extent, they differ in methodology: Sirius relies solely on SFT and integrates incorrect trajectories via self-correction into the training dataset, whereas MALT adopts DPO, naturally utilising negative samples. These methods provide early evidence of the potential for self-improvement in multi-agent systems, though they are primarily applied in relatively simple settings (\textit{e.g.}, multi-agent debate or “generator-verifier-answerer” system). Moving forward, MaPoRL~\citep{park2025maporl} introduces task-specific reward shaping to explicitly incentivise inter-agent communication and cooperation through reinforcement learning. MARFT~\citep{liao2025marft} establishes a comprehensive bridge between conventional multi-agent reinforcement learning (MARL) and LLM-based multi-agent reinforcement tuning. Building on this, MARTI~\citep{liao2025marft} proposes a more customizable framework for reinforced multi-agent fine-tuning, supporting flexible design of both agentic structures and reward functions. Empirical results show that LLM backbones exhibit considerable improvements in cooperative capabilities during their cooperative training.

\paragraph{{Collaboration-oriented Optimisation.}} Beyond reasoning, a smaller body of work focuses on enhancing communication and collaboration abilities within multi-agent systems. The core assumption is that LLM agents are not inherently effective team players, and their collaborative communication skills require targeted training. An early example is COPPER~\citep{bo2024reflective}, which employs PPO to train a shared reflector that generates high-quality, role-aware personalised reflections for multi-agent collaboration trajectories. OPTIMA~\citep{chen2024optima} more directly targets communication efficiency in multi-agent systems (measured by token usage and communication readability) and explores achieving an effectiveness-efficiency trade-off via SFT, DPO, and hybrid methods. It reports a 2.8× performance gain with less than 10\% of the token cost on tasks demanding intensive information exchange, which vividly demonstrates the promising potential of scaling agents' collaborative capabilities. Further, MaPoRL~\citep{park2025maporl} argues that the prevalent paradigm of prompting out-of-the-box LLMs and relying solely on their innate collaborative abilities is questionable. Instead, it introduces carefully designed reinforcement learning signals within a multi-agent debate framework to explicitly elicit collaborative behaviours, encouraging agents to communicate more frequently and with higher quality. 

\section{Domain-Specific Optimisation}
\label{sec:domain-specific_optimisation}

While previous sections have focused on agent optimisation and evolution techniques in general-domain settings, domain-specific agent systems introduce unique challenges that require tailored optimisation strategies. These domains, such as biomedicine~\citep{almansoori2025self}, programming~\citep{tang2024codeagent}, scientific research~\citep{pu2025piflow}, game-playing~\citep{belle2025agents}, computer use~\citep{sun2025seagent}, and finance \& legal research, are often characterised by specialised task structures, domain-specific knowledge bases, distinct data modalities, and operational constraints. Such factors can significantly influence how agents are designed, optimised, and evolved. In this section, we survey recent advances in domain-specific agent optimisation and evolution, highlighting effective techniques that have been developed to meet the unique demands of each domain.

\subsection{Domain-Specific Optimisation in Biomedicine} 
In the biomedical domain, agent optimisation focuses on aligning agent behaviours with the procedural and operational requirements of real-world clinical settings. Recent studies have demonstrated the effectiveness of domain-specific agent design in two key application areas: medical diagnosis~\citep{donner2018solving,almansoori2025self,zhuang2025learning} and molecular discovery~\citep{m2024augmenting,inoue2025drugagent}. In what follows, we examine representative agent optimisation strategies within these two domains.

\subsubsection{Medical Diagnosis}

Medical diagnosis requires determining a patient’s condition based on clinical information such as symptoms, medical history, and diagnostic test results~\citep{KONONENKO200189,donner2018solving}. Recent research has increasingly explored the use of autonomous agents in this context, enabling systems to automatically conduct diagnostic dialogues, pose clarifying questions, and generate plausible diagnostic hypotheses~\citep{li2024agent,chen2025enhancing,zuo2025kg4diagnosis,ghezloo2025pathfinder}. These agents often operate under uncertain conditions, making decisions based on incomplete or ambiguous patient information~\citep{chen2025enhancing}. The diagnostic process typically involves multi-turn interactions, during which agents elicit missing information through follow-up enquiries~\citep{chen2025enhancing}. Moreover, to support robust clinical reasoning, agents often require integrating external knowledge bases or interacting with specialised medical tools for information retrieval and evidence-based reasoning~\citep{feng2025m,fallahpour2025medrax}.

Given these domain-specific requirements, recent studies have focused on developing agent architectures specifically optimised for medical diagnosis~\citep{li2024mmedagent,almansoori2025self,ghezloo2025pathfinder, wang2025medagent}. One promising research direction focuses on multi-agent systems, which have shown strong potential for modelling the complexity and multi-step reasoning involved in medical diagnosis. These approaches can be broadly classified into two categories: \textit{simulation-driven} and \textit{collaborative designs}. Simulation-driven systems aim to reproduce real clinical settings by assigning specific roles to agents and enabling them to learn diagnostic strategies through interactions within a simulated medical environment.
For instance, MedAgentSim~\citep{almansoori2025self} introduces a self-evolving simulation framework that integrates experience replay, chain-of-thought ensembling, and CLIP-based semantic memory to support diagnostic reasoning. PathFinder~\citep{ghezloo2025pathfinder} targets histopathological analysis by orchestrating multiple agents to emulate expert diagnostic workflows on gigapixel-scale medical images. In contrast, collaborative multi-agent systems focus on collective decision-making and collaboration among agents. For example, MDAgents~\citep{kim2024mdagents} enables adaptive collaboration among multiple agents, where a moderator agent is responsible for integrating diverse suggestions and consulting external knowledge sources as needed. MDTeamGPT~\citep{chen2025mdteamgpt} extends this paradigm to multidisciplinary consultation, supporting self-evolving, team-based diagnostic processes through reflective discussion mechanisms. 

Another line of work on agent optimisation for diagnosis focuses on tool integration and multimodal reasoning. For instance, MMedAgent~\citep{li2024mmedagent} addresses the generalisability limitations of existing multimodal LLMs by dynamically incorporating specialised medical tools across different modalities. To improve clinical reliability, MedAgent-Pro~\citep{wang2025medagent} introduces diagnostic planning guided by established clinical criteria and integrates multimodal evidence via task-specific tool agents. In contrast to fixed agent architectures, recent work has explored more flexible designs that adapt based on diagnostic performance. For example, \cite{zhuang2025learning} proposes a graph-based agent framework where the reasoning process is continuously adjusted using feedback from diagnostic results. 
These approaches highlight specialisation, multimodality, and interactive reasoning as key principles for developing agent-based systems in medical diagnosis. 

\subsubsection{Molecular Discovery and Symbolic Reasoning}

Molecular discovery within biomedical domains demands precise symbolic reasoning over chemical structures, reaction pathways, and pharmacological constraints~\citep{bilodeau2022generative,makke2024interpretable,m2024augmenting}. To support molecular discovery, recent agent-based systems have introduced tailored techniques such as integrating chemical analysis tools, enhancing memory for 
knowledge retention, and enabling multi-agent collaboration~\citep{mcnaughton2024cactus,inoue2025drugagent}. One key approach is domain-specific tool integration, which allows agents to perform chemical reasoning through interaction with executable chemical operations.
For instance, CACTUS~\citep{mcnaughton2024cactus} equips agents with cheminformatics tools such as RDKit~\citep{landrum2013rdkit} to ensure the generation of chemically valid outputs. By grounding reasoning in domain-specific toolsets, CACTUS achieves significantly better performance than agents without tool integration. Similarly, LLM-RDF~\citep{m2024augmenting} automates chemical synthesis by coordinating specialised agents, each responsible for a specific task and equipped with corresponding tools for literature mining, synthesis planning, or reaction optimisation.

Another prominent line of research leverages memory-enabled reasoning~\citep{hu2025osda,inoue2025drugagent}, where agents learn from prior experience by recording how previous problems were solved. ChemAgent~\citep{tang2025chemagent} breaks down complex chemical tasks into smaller subtasks, which are stored within a structured memory module, enabling efficient retrieval and refinement. OSDA Agent~\citep{hu2025osda} extends this approach by introducing a self-reflective mechanism, where failed molecule proposals are abstracted into structured memory updates that inform and enhance future decision-making. In parallel, multi-agent coordination provides additional benefits. DrugAgent~\citep{inoue2025drugagent} introduces a coordinator architecture that integrates evidence from machine learning-based predictors, biomedical knowledge graphs, and literature search agents. It employs Chain-of-Thought and ReAct~\citep{yao2023react} frameworks to support interpretable, multi-source reasoning. LIDDIA~\citep{averly2025liddia} generalises this design by assigning modular roles, i.e., reasoner, executor, evaluator, and memory, which collectively emulate iterative workflows in medicinal chemistry and facilitate multi-objective molecule evaluation.

\subsection{Domain-Specific Optimisation in Programming}

In the programming domain, agent optimisation focuses on aligning agent behaviours with the procedural and operational requirements of established software engineering workflows. Recent studies have demonstrated the effectiveness of domain-specific agent design in two key application areas: code refinement~\citep{rasheed2024codepori,tang2024codeagent,pan2025codecor} and code debugging~\citep{lee2024unifieddebuggingapproachllmbased,puvvadi2025coding,adnan2025large}. In what follows, we examine representative agent optimisation strategies within these two domains.

\subsubsection{Code Refinement}

Code refinement involves the iterative improvement of code quality, structure, and correctness while preserving its original functionality~\citep{yang2024enhancing,he2025llm,islam2025codesim}. Recent studies have increasingly investigated agent-based systems that support domain-specific optimisation for this task, focusing on self-improvement, collaborative workflows, and integration with programming tools~\citep{madaan2023self,tang2024codeagent,rahman2025marco}. These systems are designed to emulate human-in-the-loop refinement processes, enforce adherence to software engineering best practices, and ensure that code remains robust, readable, and maintainable throughout iterative development cycles. One critical optimisation strategy involves self-feedback mechanisms, where agents critique and revise their own outputs. For example, Self-Refine~\citep{madaan2023self} introduces a lightweight framework in which a language model generates natural language feedback on its own outputs and subsequently revises the code accordingly. Similarly, CodeCriticBench~\citep{zhang2025codecriticbench} presents a comprehensive benchmark designed to assess the self-critiquing and refinement capabilities of LLMs, where agents are evaluated on their ability to identify, explain, and revise code defects through structured natural language feedback. LLM-Surgeon~\citep{van2023llm} proposes a systematic framework in which a language model diagnoses structural and semantic issues within its own code outputs and applies targeted edits based on learned repair patterns, thereby optimising code quality while preserving functionality. These approaches eliminate the need for task-specific retraining, providing consistent improvements in code quality.

Another line of research explores experience-driven learning, where agents improve their problem-solving capabilities by relying on memory-enabled reasoning, systematically recording and reusing solutions to previously encountered tasks~\citep{wang2024openhands,tang2024codeagent,pan2025codecor}. For example, AgentCoder~\citep{huang2023agentcoder} and CodeAgent~\citep{tang2024codeagent} simulate collaborative development workflows by assigning specialised roles to individual agents, such as coder, reviewer, and tester, which iteratively improve code through structured dialogue cycles. These systems support collective evaluation and revision, promoting role specialisation and deliberative decision-making. Additionally, tool-enhanced frameworks such as CodeCoR~\citep{pan2025codecor} and OpenHands~\citep{wang2024openhands} incorporate external tools and modular agent interactions to facilitate dynamic code pruning, patch generation, and context-aware refinement. VFlow~\citep{wei2025vflow} reformulates the workflow optimisation problem of Verilog code generation task as a search task on a graph of LLM nodes with code-based representations, employing a Cooperative Evolution with Past Experience MCTS (CEPE-MCTS) algorithm. These developments highlight iterative feedback, modular design, and interactive reasoning as essential principles for building adaptive agent-based systems for code refinement.

\subsubsection{Code Debugging}

Code debugging presents intricate challenges that require precise fault localisation, execution-aware reasoning, and iterative correction. These capabilities are typically absent in general-purpose LLMs~\citep{puvvadi2025coding, mannadiar2010debugging}. To address these challenges, domain-specific optimisation focuses on aligning agent roles and workflows with the structured reasoning patterns and tool usage observed in human debugging practices. A key strategy involves leveraging runtime feedback to facilitate self-correction. For example, Self-Debugging~\citep{chen2023teaching} and Self-Edit~\citep{zhang2023self} exemplify this approach by incorporating execution traces into the debugging process. These agents operate through internal cycles of fault identification, natural language-based reasoning, and targeted code revision, enabling autonomous debugging without external supervision.

Recent research has explored modular agent architectures specifically designed to support the multi-stage structure of debugging workflows. For instance, PyCapsule~\citep{adnan2025large} introduces a separation of responsibilities between a programmer agent and an executor agent, thereby distinguishing code generation from semantic validation. More advanced systems, including Self-Collaboration~\citep{dong2024self} and RGD~\citep{jin2024rgd}, employ collaborative pipelines in which agents are assigned specialised roles such as tester, reviewer, or feedback analyser, mirroring professional debugging practices. Additionally, FixAgent~\citep{lee2024unifieddebuggingapproachllmbased} extends this paradigm through hierarchical agent activation, dynamically dispatching different agents based on bug complexity and required depth of analysis.

\subsection{Domain-Specific Optimisation in Financial and Legal Research}

In financial and legal domains, agent optimisation focuses on tailoring multi-agent architectures, reasoning strategies, and tool integration to the procedural and operational demands of domain-specific workflows~\citep{sun2024lawluo,he2024agentscourt,li2025hedgeagents}. Recent studies have demonstrated the effectiveness of such domain-specific designs in two key application areas: financial decision-making~\citep{li2023tradinggpt,yu2024fincon,wang2024peer} and legal reasoning~\citep{di2023multi,chen2024agentcourt}, where modular design, collaborative interaction, and rule-grounded reasoning are essential for reliable performance. In what follows, we examine representative agent optimisation strategies within these two domains.

\subsubsection{Financial Decision-Making}

Financial decision-making requires agents to operate under uncertain and rapidly changing conditions, reason over volatile market dynamics, and integrate heterogeneous information sources such as numerical indicators, news sentiment, and expert knowledge~\citep{li2023tradinggpt,sarin2024unleashing,chudziak2025elliottagents}. In response to these domain-specific demands, recent research has focused on developing multi-agent architectures tailored to the procedural and cognitive requirements of financial environments~\citep{fatemi2024finvision,luo2025llm}. One critical strategy involves conceptual and collaborative agent design. For instance, FinCon~\citep{yu2024fincon} proposes a synthesised multi-agent system built on LLMs, employing conceptual verbal reinforcement and domain-adaptive fine-tuning to enhance decision stability and policy alignment in dynamic markets. PEER~\citep{wang2024peer} extends this paradigm through a modular agent architecture comprising expert, retriever, and controller roles, which interact under a unified tuning mechanism to balance task specialisation with general adaptability. FinRobot~\citep{yang2024finrobot} further advances this line of work by integrating external tools for model-grounded reasoning, enabling agents to connect high-level strategies with executable financial models and real-time data streams.

Another line of work on agent optimisation for financial decision-making focuses on sentiment analysis and reporting~\citep{xing2024designing,tian2025template,raza2025trism}. Heterogeneous LLM agent architectures~\citep{xing2024designing} enhance robustness in financial reporting by combining specialised sentiment modules with rule-based validators to ensure compliance with domain-specific guidelines. Similarly, template-based reporting frameworks~\citep{tian2025template} decompose report generation into agent-driven retrieval, validation, and synthesis stages, enabling iterative refinement through real-world feedback. These approaches demonstrate the potential of self-evolving multi-agent systems to provide reliable, interpretable, and context-aware decision support in complex financial environments.

\subsubsection{Legal Reasoning}

Legal reasoning requires agents to interpret structured legal rules, analyse case-specific evidence, and produce outputs that are consistent with institutional regulations and judicial standards~\citep{xu2023multi,yuan2024can,jiang2025agentsbench}. To address these domain-specific demands, recent research has explored multi-agent systems tailored to the procedural and interpretive requirements of legal settings~\citep{di2023multi,hu2023language,chen2024agentcourt}. One significant direction involves collaborative agent frameworks that simulate judicial processes and support structured argumentation. For instance, LawLuo~\citep{sun2024lawluo} introduces a co-run multiagent architecture in which legal agents are assigned specialised roles such as document drafting, legal argument generation, and compliance validation, all operating under the supervision of a central controller to ensure procedural consistency and legal correctness. Multi-Agent Justice Simulation~\citep{di2023multi} and AgentCourt~\citep{chen2024agentcourt} extend this paradigm to model adversarial trial procedures, enabling agents to participate in role-based interactions that emulate real-world courtroom dynamics. In particular, AgentCourt incorporates self-evolving lawyer agents that refine their strategies through reflective self-play, leading to improved debate quality and enhanced procedural realism.

Another line of work focuses on structured legal reasoning and domain-grounded interpretability. LegalGPT~\citep{shi2024legalgpt} integrates a legal chain-of-thought framework within a multi-agent system, guiding legal reasoning through interpretable and rule-aligned steps. Similarly, AgentsCourt~\citep{he2024agentscourt} combines courtroom debate simulation with legal knowledge augmentation, enabling agents to perform judicial decision-making grounded in codified rules and case precedents. These approaches highlight the importance of rule grounding, modular role design, and collaborative reasoning in the development of robust, transparent, and legally reliable agent systems.

\section{Evaluation}
\label{sec:evaluation}

The rapid emergence of autonomous LLM-based agents has underscored the need for rigorous, multidimensional evaluation frameworks. As these agents are deployed across increasingly diverse tasks and environments, recent research has introduced a range of benchmarks and methodologies to assess not only task completion but also reasoning quality, generalisation ability, and compliance with safety and alignment standards. Evaluation is no longer viewed as a static endpoint but as a dynamic feedback mechanism: fine-grained performance signals are now used to guide agent optimisation, prompt refinement, and dataset augmentation, enabling self-evolving systems that continuously acquire new capabilities and address failure cases. Current evaluation paradigms encompass structured benchmark tasks with standardised metrics, safety- and alignment-oriented audits, and LLM-as-a-judge approaches that leverage large models as flexible, scalable evaluators. 

\subsection{Benchmark-based Evaluation}

\subsubsection{Tool and API-Driven Agents}
Tool-augmented agents are evaluated based on their ability to invoke external APIs and functions to solve problems that exceed the scope of their intrinsic knowledge. Benchmarks such as ToolBench~\citep{xu2023tool}, API-Bank~\citep{li2023api}, MetaTool~\citep{huang2023metatool}, and ToolQA~\citep{zhuang2023toolqa} define tasks that require tool usage and assess both the correctness and efficiency of API calls. Many of these evaluations employ simulated APIs or sandboxed environments, measuring task success alongside interaction efficiency. Early studies have shown that agents often overfit to specific tool schemas, exhibiting limited generalisation to previously unseen APIs. To address this limitation, recent benchmarks such as GTA~\citep{wang2024gta} and AppWorld~\citep{trivedi2024appworld} introduce more realistic, multi-step tasks that require planning and coordination across multiple tools, while placing greater emphasis on process-oriented evaluation metrics. This trend reflects a broader shift towards richer, reasoning-aware evaluations that assess not only final outcomes but also the quality of the decision-making process.

\subsubsection{Web Navigation and Browsing Agents}
Web agents are evaluated on their ability to interact with websites, extract information, and complete real-world online tasks. Benchmarks such as BrowseComp~\citep{wei2025browsecomp}, MM-BrosweComp~\citep{li2025mm}, WebArena~\citep{zhou2023webarena}, VisualWebArena~\citep{koh2024visualwebarena}, WebCanvas~\citep{pan2024webcanvas}, WebWalker~\citep{wu2025webwalker}, and AgentBench~\citep{liu2023agentbench} have progressively increased the realism and diversity of web-based evaluations, spanning simulated and live environments. These benchmarks test navigation skills, adaptability to interface changes, and the integration of textual and visual information. Recent work incorporates intermediate metrics (e.g., sub-goal completion) and robustness assessments, though reproducibility and generalisation remain challenging due to the dynamic nature of the web.

\subsubsection{Multi-Agent Collaboration and Generalists}
As agents become more general-purpose, new benchmarks target multi-agent coordination and cross-domain competence. MultiAgentBench~\citep{zhu2025multiagentbench} and SwarmBench~\citep{ruan2025benchmarking} evaluate collaboration, competition, and decentralised coordination among LLM agents, assessing both task completion and the quality of communication and strategy. Generalist benchmarks such as GAIA~\citep{mialon2023gaia} and AgentBench~\citep{liu2023agentbench} test adaptability across diverse environments, from web navigation to coding and database queries. Recent work, \cite{wang2025efficientagentsbuildingeffective} further explores the GAIA benchmark to analyse the efficiency–effectiveness trade-off in agentic systems, proposing \textsc{Efficient Agents}, a framework that achieves competitive performance with significantly reduced operational costs. These evaluations highlight challenges in aggregating metrics across heterogeneous tasks, risks of overfitting to narrow scenarios, and the need for unified, holistic leaderboards.

\subsubsection{GUI and Multimodal Environment Agents}
GUI and multimodal benchmarks challenge agents to operate in rich, interactive environments that combine textual and visual inputs. Mobile-Bench~\citep{deng2024mobile}, AndroidWorld~\citep{rawles2024androidworld}, CRAB~\citep{xu2024crab}, GUI-World~\citep{chen2024gui}, and OSWorld~\citep{xie2024osworld} simulate realistic apps and operating systems, requiring complex action sequences. Tasks often combine natural language understanding, visual perception, and API invocation. Evaluations measure task success, state management, perception accuracy, and adaptability to GUI changes. However, the diversity of GUI environments makes standardisation and reproducibility difficult, and agents remain brittle when faced with interface variability.

\subsubsection{Domain-Specific Task Agents}
Domain-focused benchmarks in coding (SWE-bench~\citep{jimenez2023swe}), data science (DataSciBench~\citep{zhang2025datascibench}, MLGym~\citep{nathani2025mlgym}), enterprise productivity (WorkBench~\citep{styles2024workbench}), and scientific research (OpenAGI~\citep{ge2023openagi}, SUPER~\citep{bogin2024super}) assess specialised competencies that integrate planning, tool use, and adherence to domain norms. SWE-bench, for example, evaluates code-editing agents on real GitHub repositories, while AgentClinic~\citep{schmidgall2024agentclinic} and MMedAgent~\citep{li2024mmedagent} test multimodal reasoning in clinical settings. 
Evaluation criteria have expanded from binary success measures to encompass metrics such as test pass rates, policy adherence, and conformity to domain-specific constraints. Despite these advances, inconsistencies in metric definitions and persistent gaps in generalisation remain significant challenges.

\subsection{LLM-based Evaluation}

\subsubsection{LLM-as-a-Judge}
The LLM-as-a-Judge paradigm refers to employing large language models to assess the quality of outputs generated by AI systems, such as text, code, or conversational responses, via structured prompts~\citep{arabzadeh2024towards,li2024llms,qian2025enhancing}. This approach has attracted attention as a scalable and cost-effective alternative to conventional evaluation methods, including human judgment and automatic metrics (e.g., BLEU, ROUGE), which often fail to capture semantic depth or coherence~\citep{arabzadeh2024towards}. LLM judges typically operate in two modes: \textit{pointwise} evaluation~\citep{ruan2024liveideabench}, where outputs are scored directly against criteria such as factuality and helpfulness, and \textit{pairwise} comparison, where two outputs are compared and the preferred one is selected with justification~\citep{li2024llms,zhao2024auto}.

Recent studies demonstrate that LLM-based evaluations can correlate with human judgments, in some cases reaching parity with inter-annotator agreement levels~\citep{arabzadeh2024towards}. However, these methods are sensitive to prompt design and susceptible to biases introduced by subtle instructional variations~\citep{arabzadeh2024towards,zhao2024auto}. Furthermore, single-step, output-focused evaluations may overlook the reasoning depth in multi-step processes~\citep{zhuge2024agent,wang2025dynamically}. To address these limitations, enhancements have been proposed, including multi-agent deliberation frameworks such as CollabEval~\citep{qian2025enhancing} and structured meta-evaluation benchmarks to calibrate and improve the reliability of LLM judges~\citep{li2024llms,zhao2024auto}.

\subsubsection{Agent-as-a-Judge}
The Agent-as-a-Judge framework extends LLM-based evaluation by employing full-fledged agentic systems capable of multi-step reasoning, state management, and tool use to critique other AI agents~\citep{zhuge2024agent,zhao2024auto,qian2025enhancing}. Different from traditional LLM judges, which focus solely on final outputs, agent judges evaluate the entire reasoning trajectory, capturing decision-making processes and intermediate actions~\citep{zhuge2024agent}. For example, \cite{zhuge2024agent} applied an agent judge to the DevAI benchmark for code-generation agents. The framework incorporated specialised modules to analyse intermediate artefacts, construct reasoning graphs, and validate hierarchical requirements, resulting in evaluations that aligned more closely with human expert judgments than traditional LLM-based approaches. Agent judges also delivered substantial efficiency gains, reducing evaluation time and cost relative to manual review~\citep{zhuge2024agent,zhao2024auto}.

Nevertheless, implementing the Agent-as-a-Judge methodology introduces additional complexity and raises challenges for generalisation to domains other than code generation. Current research seeks to improve adaptability and simplify deployment across a broader range of AI tasks~\citep{zhao2024auto,qian2025enhancing}.

\subsection{Safety, Alignment, and Robustness in Lifelong Self-Evolving Agents}
In the context of the \ThreeLaws{}, \textbf{Endure}, the maintenance of safety and stability during any modification, forms the primary constraint on all other forms of adaptation. For lifelong, self-evolving agentic systems, safety is not a one-off certification but an ongoing requirement: every evolution step, from prompt updates to topology changes, must be assessed for unintended or malicious behaviours. This necessitates evaluation protocols that are \emph{continuous}, \emph{granular}, and \emph{scalable}, ensuring that agents can remain aligned while adapting over extended lifetimes.

Recent work has introduced diverse evaluation paradigms. Risk-focused benchmarks such as \textsc{AgentHarm}~\citep{andriushchenko2024agentharm} measure an agent’s propensity to comply with explicitly malicious multi-step requests—requiring coherent tool use to execute harmful objectives such as fraud or cybercrime, revealing that even leading LLMs can be coaxed into complex unsafe behaviours with minimal prompting. Domain-specific probes such as \textsc{RedCode}~\citep{guo2024redcode} (code security) and \textsc{MobileSafetyBench}~\citep{lee2024mobilesafetybench} (mobile control) stress-test agents in realistic, sandboxed environments. Behavioural probes like \textsc{MACHIAVELLI}~\citep{pan2023rewards} explore whether agents develop unethical, power-seeking strategies under reward optimisation, highlighting the interplay between \textbf{Endure} and \textbf{Excel}, safe adaptation must not degrade core task competence.

Meta-evaluation approaches, e.g., \textsc{Agent-as-a-Judge}~\citep{zhuge2024agent}, \textsc{AgentEval}~\citep{arabzadeh2024towards}, and \textsc{R-Judge}~\citep{yuan2024r} -- position LLM agents themselves as evaluators or safety monitors, offering scalable oversight but also exposing the limitations of current “risk awareness.” These studies underline the multi-dimensional nature of safety, where accuracy alone is insufficient; over-reliance on correctness metrics can conceal epistemic risks and systemic biases~\citep{li2025accuracy}. Legal alignment tests such as \textsc{SafeLawBench}~\citep{cao_safelawbench_2025} further show that even state-of-the-art models struggle to satisfy established legal principles, reflecting the difficulty of codifying alignment in domains with open-textured norms.

Despite these advances, most current evaluations are \textit{snapshot-based}, assessing agents at a single point in time. For MASE systems, safety evaluation must itself become dynamic -- continuously monitoring, diagnosing, and correcting behaviours as the system evolves. Developing longitudinal, evolution-aware benchmarks that track safety, alignment, and robustness across the full lifecycle of an agent ecosystem remains an open and urgent challenge.

\section{Challenges and Future Directions}
\label{sec:challenges_and_directions}

Despite rapid advances, the evolution and optimisation of AI agents still face fundamental obstacles. These challenges are closely tied to the \ThreeLaws{} and need to be addressed to realise the vision of lifelong agentic systems. We group the key open problems accordingly.
\subsection{Challenges}
\subsubsection{Endure – Safety Adaptation}
\begin{enumerate}[label=(\arabic*)]
    \item \textbf{Safety, Regulation, and Alignment.} Most optimisation pipelines prioritise task metrics over safety constraints, neglecting risks such as unintended behaviours, privacy breaches, and misaligned objectives. The dynamic nature of evolving agents undermines existing legal frameworks (e.g., EU AI Act, GDPR), which assume static models and fixed decision logic. This calls for new evolution-aware audit mechanisms, adaptive licences, provable-safety sandboxes, and legal protocols capable of tracking and constraining an agent’s self-directed evolutionary path.
    \item \textbf{Reward Modelling and Optimisation Instability.} Learned reward models for intermediate reasoning steps often suffer from dataset scarcity, noisy supervision, and feedback inconsistency, leading to unstable or divergent agent behaviours. Stability is central to safety: even small perturbations in inputs or update rules can undermine the trustworthiness of an evolving workflow.
\end{enumerate}

\subsubsection{Excel – Performance Preservation}
\begin{enumerate}[label=(\arabic*)]
    \item \textbf{Evaluation in Scientific and Domain-Specific Scenarios.} In domains like biomedicine or law, reliable ground truth is often absent or disputed, complicating the construction of trustworthy feedback signals for optimisation.
    \item \textbf{Balancing Efficiency and Effectiveness in MAS Optimisation.} Large-scale multi-agent optimisation improves task performance but incurs significant computational cost, latency, and instability. Designing methods that explicitly trade off effectiveness against efficiency remains unresolved.
    \item \textbf{Transferability of Optimised Prompts and Topologies.} Optimised prompts or agent topologies are often brittle, showing poor generalisation across LLM backbones with differing reasoning abilities. This undermines scalability and reusability in production settings.
\end{enumerate}

\subsubsection{Evolve – Autonomous Optimisation}
\begin{enumerate}[label=(\arabic*)]
    \item \textbf{Optimisation in Multimodal and Spatial Environments.} Most optimisation algorithms are text-only, yet real-world agents must process multimodal inputs and reason in spatially grounded or continuous environments. This demands internal world models and perceptual–temporal reasoning.
    \item \textbf{Tool Use and Creation.} Current methods typically assume a fixed toolset, overlooking the autonomous discovery, adaptation, and co-evolution of tools alongside agents.
\end{enumerate}

\subsection{Future Directions}

Looking forward, many of these limitations point to promising research avenues. We highlight several directions and link them to their role in the \textbf{MOP→MOA→MAO→MASE} paradigm shift.

\begin{enumerate}[label=(\arabic*)]
    \item \textbf{Simulated Environments for Fully Autonomous Self-Evolution (MASE).} Develop open-ended, interactive simulation platforms where agents can iteratively interact, receive feedback, and refine prompts, memory, tools, and workflows via closed-loop optimisation.
    \item \textbf{Advancing Tool Use and Creation (MAO$\rightarrow$MASE).} Move beyond static tool invocation toward agents that adaptively select, compose, or create tools. Incorporate reinforcement learning and feedback-driven strategies, paired with robust evaluation pipelines.
    \item \textbf{Real-World Evaluation and Benchmarking (Cross-stage).} Create benchmarks and protocols that reflect real-world complexity, support interaction-based and longitudinal assessment, and align with long-term improvement signals.
    \item \textbf{Effectiveness–Efficiency Trade-offs in MAS Optimisation (MAO).} Design optimisation algorithms that jointly model performance and resource constraints, enabling MAS deployment under strict latency, cost, or energy budgets.
    \item \textbf{Domain-Aware Evolution for Scientific and Specialised Applications (MASE).} Tailor evolution methods to domain-specific constraints in science, medicine, law, or education, integrating heterogeneous knowledge sources, bespoke evaluation criteria, and regulatory compliance.
\end{enumerate}

\noindent
\textbf{Outlook.} Addressing these challenges will require optimisation pipelines that are not only high-performing and domain-adaptive, but also safe, regulation-aware, and self-sustaining. Embedding these solutions within the \textbf{MOP→MOA→MAO→MASE} trajectory, and grounding them in the \ThreeLaws{}, offers a coherent roadmap toward truly \textit{lifelong, autonomous agentic systems} -- systems that can \textbf{endure}, \textbf{excel}, and \textbf{evolve} across the full span of their operational lifetimes.

\section{Conclusions}
\label{sec:conclusions}
In this survey, we have presented a comprehensive overview of the emerging paradigm of self-evolving AI agents, which bridge the static capabilities of foundation models with the continuous adaptability required by lifelong agentic systems. We situated this evolution within a unified four-stage trajectory, from Model Offline Pretraining (MOP) and Model Online Adaptation (MOA), through Multi-Agent Orchestration (MAO), and ultimately to Multi-Agent Self-Evolving (MASE), highlighting the progressive shift from static, human-configured models to dynamic, autonomous ecosystems.

To formalise this transition, we introduced a conceptual framework that abstracts the feedback loop underlying agent evolution, with four key components: Inputs, Agent System, Objectives, and Optimisers, that together determine how agents improve through continual interaction with their environment. Building on this, we systematically reviewed optimisation techniques across agent components, domain-specific strategies, and evaluation methodologies critical for building adaptive and resilient agentic systems.

We also proposed the \ThreeLaws{}, Endure (safety adaptation), Excel (performance preservation), and Evolve (autonomous evolution), as guiding principles to ensure that lifelong self-improvement remains safe, effective, and aligned. These laws are not mere principles but practical design constraints, ensuring that the path toward autonomy remains aligned with safety, performance, and adaptability. They serve as the guardrails for the MASE paradigm, guiding research from narrow, single-shot optimisation toward continuous, open-ended self-improvement.

Looking forward, the ability to endure, excel, and evolve will be decisive for agents operating in dynamic, real-world environments, whether in scientific discovery, software engineering, or human–AI collaboration. Achieving this will demand breakthroughs in scalable optimisation algorithms, lifelong evaluation protocols, safe coordination in heterogeneous agent environments, and mechanisms for adapting to unforeseen domains.

We hope this survey serves as both a reference point and a call to action to build an ecosystem of self-evolving AI agents that do not simply execute tasks, but live, learn, and last. By aligning technical innovation with principled self-evolution, we can pave the way toward truly autonomous, resilient, and trustworthy lifelong agentic systems.

\section*{Acknowledgements} 
We would like to thank Shuyu Guo for his valuable contributions to the early-stage exploration and literature review on agent optimisation.

\bibliographystyle{assets/plainnat}
\bibliography{citation}
\clearpage
\end{document}